\pdfoutput=1
\documentclass[11pt]{article}

\usepackage{EMNLP2023}

\usepackage{times}
\usepackage{latexsym}
\usepackage[T1]{fontenc}
\usepackage[utf8]{inputenc}
\usepackage{microtype}
\usepackage{inconsolata}
\usepackage{subcaption}
\usepackage{booktabs}
\usepackage{bm} 
\usepackage[inline]{enumitem}
\usepackage{acronym}
\usepackage[skip=1pt]{caption}
\usepackage{algorithm}
\usepackage{algpseudocode}
\usepackage{CJKutf8}
\usepackage{graphicx}
\usepackage{amsmath}
\usepackage{multirow}
\usepackage{amsfonts}
\usepackage{colortbl}
\AtBeginDocument{%
  }

\newcommand{\highlight}[1]{\textcolor{darkblue}{#1}}
\newcommand{\pemph}[1]{\emph{\textbf{#1}}}

\acrodef{CRS}{conversational recommender system}
\acrodef{ECR}{E-commerce conversational recommendation}
\acrodef{LLM}{large language model}
\acrodef{UNECR}{user needs-centric E-commerce conversational recommendation}
\acrodef{CRD}{conversational recommendation dataset}
\acrodef{U-NEED}{user needs-centric E-commerce conversational recommendation dataset}
\acrodef{TOD}{task-oriented dialogue}
\acrodef{TODS}{task-oriented dialogue system}
\acrodef{ODD}{open-domain dialogue}
\acrodef{ODDS}{open-domain dialogue system}
\acrodef{KB}{knowledge base}
\acrodef{KG}{knowledge graph}
\acrodef{BehaviorCRS}{user behavior-aware conversational recommender system}
\acrodef{PLM}{pre-trained language model}

\newcommand{\header}[1]{\vspace{0.5mm}\noindent\textbf{#1.}}

\definecolor{mygray}{gray}{.92}

\title{Conversational Recommender System and Large Language Model \\ Are Made for Each Other in E-commerce Pre-sales Dialogue}
\author{Yuanxing Liu$^1$\quad
Wei-Nan Zhang$^1$\footnotemark[2]\quad
Yifan Chen$^1$\quad
Yuchi Zhang$^1$\quad
Haopeng Bai$^1$\quad \\ \AND 
Fan Feng$^2$\quad 
Hengbin Cui$^2$\quad
Yongbin Li$^2$\quad
Wanxiang Che$^1$ \\
$^1$\normalsize{Research Center for Social Computing and Information Retrieval}\\[-.05cm]
\normalsize{Harbin Institute of Technology, China}\\[-.05cm]
$^2$\normalsize{Independent, China}\\[-.05cm]
{\small\tt\{yxliu, wnzhang, yfchen, yczhang, hpbai, car\}@ir.hit.edu.cn}\\
{\small\tt \{fengfan.blender, alexcui.chb, liyb821\}@gmail.com}}

\begin{document}
\maketitle

\renewcommand{\thefootnote}{\fnsymbol{footnote}}
\footnotetext[2]{Corresponding author.}

\renewcommand{\thefootnote}{\arabic{footnote}}
\begin{abstract}
% E-commerce pre-sales dialogue is to understand and elicit user needs and preferences to provide suitable product recommendations.
E-commerce pre-sales dialogue aims to understand and elicit user needs and preferences for the items they are seeking so as to provide appropriate recommendations.
\Acfp{CRS} learn user representation and provide accurate recommendations based on dialogue context, but rely on external knowledge.
\Acfp{LLM} generate responses that mimic pre-sales dialogues after fine-tuning, but lack domain-specific knowledge for accurate recommendations.
Intuitively, the strengths of \ac{LLM} and \ac{CRS} in E-commerce pre-sales dialogues are complementary, yet no previous work has explored this. 
This paper investigates the effectiveness of combining \ac{LLM} and \ac{CRS} in E-commerce pre-sales dialogues, proposing two collaboration methods: \emph{\ac{CRS} assisting \ac{LLM}} and \emph{\ac{LLM} assisting \ac{CRS}}. 
% To evaluate these approaches, w
We conduct extensive experiments on a real-world dataset of E-commerce pre-sales dialogues. 
% Specifically, w
We analyze the impact of two collaborative approaches with two CRSs and two LLMs on four tasks of E-commerce pre-sales dialogue.
% We conduct extensive experiments involving two \acp{CRS}, two \acp{LLM}, and two collaborative methods across four tasks on a real-world dataset of E-commerce pre-sales dialogues.
% Our findings demonstrate that the collaborations between \ac{CRS} and \ac{LLM} could yield remarkable effectiveness.
We find that collaborations between CRS and LLM can be very effective in some cases.

\end{abstract}
\section{Introduction}

% 补充为什么LLM和CRS协作的动机，补充为什么选择

\begin{figure}[!t]
    \vspace{2mm}
    \centering
    \includegraphics[width=1\linewidth]{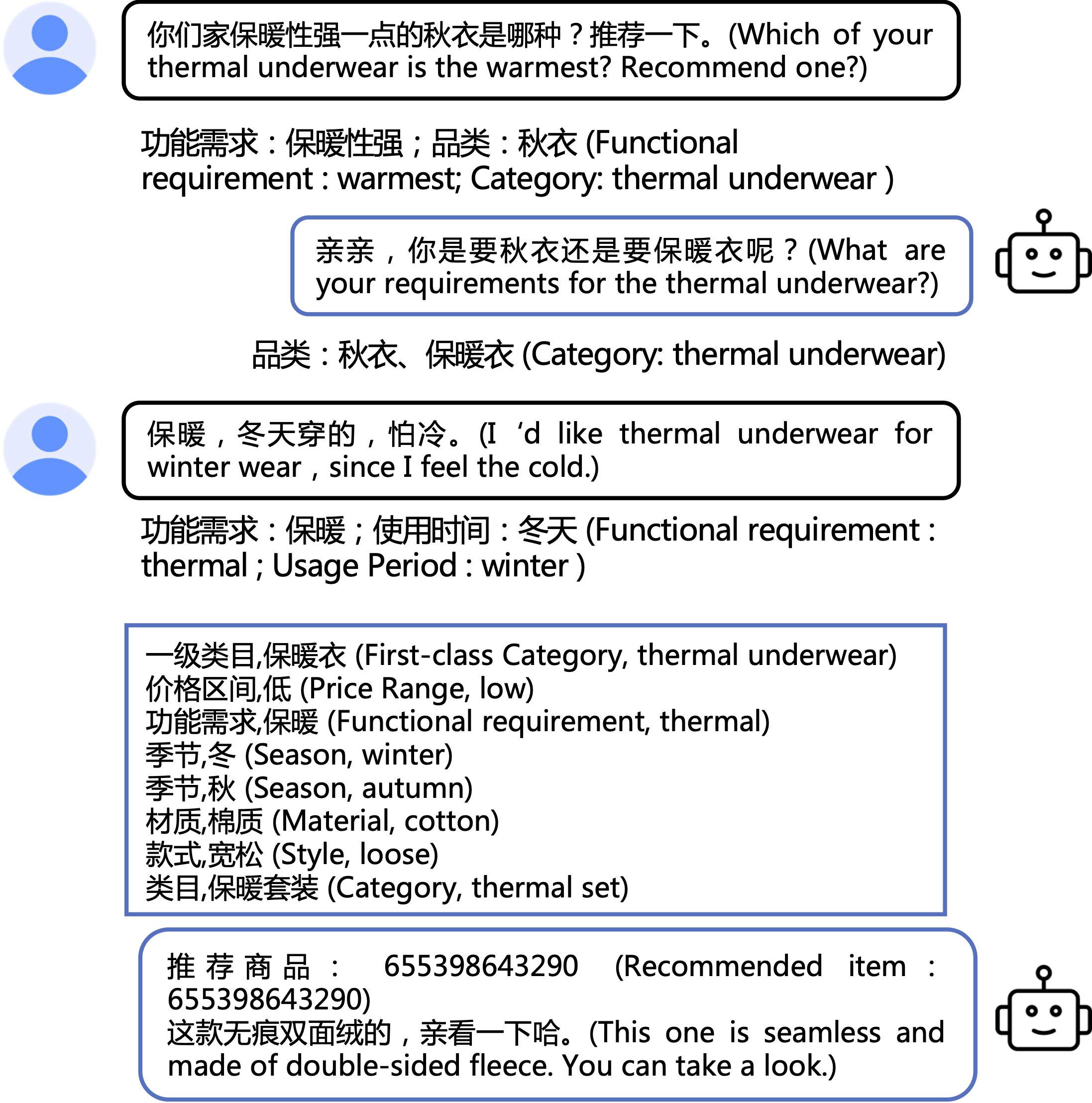}
    \caption{Example of E-commerce pre-sales dialogue. 
    % A \acf{LLM} and a \acf{CRS} have their own strengths in different tasks. Collaboration between the two can achieve better performance on all four tasks in pre-sales dialogues.
    }
    \label{fig:intro}
    % \vspace{-5mm}
\end{figure}

% 售前对话的难点
E-commerce pre-sales dialogue refers to a dialogue between a user and a customer service staff before the purchase action~\cite{chen2020jddc, zhao2021jddc}.
% Pre-sales dialogues are a very important part in E-commerce scenarios.
A high-quality pre-sales dialogue can greatly increase the purchase rate of a user.
% \footnote{}
However, there are many challenges to providing a high-quality pre-sales dialogue service~\cite{liu2023u}.
Refine.
Figure~\ref{fig:intro} shows an example of an e-commerce pre-sales dialogue.
The bot needs to interact with the user, understanding the user's needs and responding with understandable words. 
% The bot also needs to provide suitable recommendations and elicit more preferences from the user.
Additionally, it should offer appropriate recommendations and elicit further preferences from the user.

% 这个例子需要体现，LLM能够胜任的部分，以及CRS所必须的部分

% CRS的优势和劣势
% \header{Strengths and weaknesses of CRSs}
\Acfp{CRS} aim to learn relationships between user preferences and candidate product representations to provide accurate recommendations~\cite{li2018towards} and generate responses related to recommended products~\cite{liang-etal-2021-learning-neural}.
However, understanding user preferences from dialogues relies heavily on external knowledge~\cite{chen-etal-2019-towards}, e.g. DBpedia and ConceptNet.
% With external knowledge base CRS is able to know the entities appearing in the context, but still struggle to understand the semantic information in the context.
With an external knowledge base, CRS is able to recognize the entities present in the context, but it still has difficulty understanding the semantic information within the context.
% Therefore, \citet{zou2022improving} model dialogue context as item sequence, and 
%  and generate response related to recommended products~\cite{liang-etal-2021-learning-neural}.
% In addition, CRSs require external knowledge.
% For example, in the case of pre-sales dialogue understanding, \acfp{CRS} need to understand the needs expressed by the user.
% In previous work, CRSs needs to be provided with a knowledge graph in order to understand the information inside the conversation in a limited way.

% LLM的优势和劣势
% \header{Strengths and weaknesses of LLMs}
\Acfp{LLM}, which have numerous parameters pre-trained on a large amount of data, possess a wealth of knowledge that enables people to interact with them using natural language~\cite{gpt3,instructGPT}.
% LLMs, e.g. GPT-3~\cite{gpt3}, InstructGPT~\cite{instructGPT}, PaLM~\cite{palm}, Bloom~\cite{bloom}, LLaMA~\cite{llama} and GLM~\cite{du2022glm}, have attracted a lot of attention recently because of its emerging ability in natural language understanding and generation~\cite{zhao2023survey}. 
% Intuitively, LLMs has the world knowledge to understand the information in the pre-sales dialogue.
% With supervised fine-tuning tasks, LLMs can handle diverse user needs descriptions in pre-sales dialogues.
% However, LLMs do not have information about candidate products, which makes them not good at providing domain-specific recommendations.
With supervised fine-tuning tasks, LLMs can handle diverse user needs descriptions in pre-sales dialogues. However, LLMs lack information about candidate products, which makes them not suitable for providing domain-specific recommendations.

% Intuitively, with world knowledge, LLMs can handle diverse user requirements descriptions in pre-sales dialogues.
% However, to the best of our knowledge, no previous work has been done to analyze the performance of LLMs on E-commerce pre-sales dialogues.
% After training with RLHF~\cite{}, LLMs begin to have ability to chat with human on a wide range of topics~\cite{}.
% It seems that ChatGPT has perfectly addressed the memory callback in multi-turn dialogues, i.e., it is consistent with a long context~\cite{}.

% \headernodot{How do \acfp{LLM} perform in E-commerce pre-sales dialogue scenarios?}
% 协作的动机
% \header{Collaborations of CRS and LLM}
% CRSs and LLMs are made for each other in E-commerce pre-sales dialogue.
% We explore
What will happen when LLMs and CRSs collaborate?
In this paper, we explore two types of collaborations between LLMs and CRSs in E-commerce pre-sales dialogues:
% We explore two types of colloa
% , where one leads and the other assists.
\begin{enumerate*}[label=(\roman*)]
    \item {\emph{LLM assisting  CRS}.} 
    \item {\emph{CRS assisting  LLM}.} 
\end{enumerate*}
% We construct instructions for LLM based on the prediction results of CRS.
When a LLM assists a CRS, we append the generated response of the LLM to the input of the CRS. For the recommendation task, we incorporate the representation of the product predicted by the LLM into the calculation of the user representation.
When a CRS assists a LLM, we append the predictions of the CRS to the input of the LLM. For the recommendation task, we insert the recommendation list, while the other tasks insert the text.
% We design a module to encode the response of LLM and then add them to the CRS prediction probabilities.
% In this paper, we explore the effectiveness of LLM and CRS collaboration on e-commerce pre-sales conversations on a dataset of real scenarios.
% We believe that the collaboration of LLMs and CRSs could be a good solution for E-commerce pre-sales dialogue. 
% 两者协作的好处
% 两者协作的难点
% \todo{xxxxx}
% However, how to collaborate to take advantage of the strengths of both is a challenging but worthy question to explore.
% 大模型的微调和小模型的训练是不同步，两个模型如何协作是一个问题。
% 另外大模型的输出结果是回复，而小模型输出的多个候选结果。
% 如何让大模型和小模型

% 协作具体做法
% 探索了CRS与LLM协作的效果，我们
% In this paper, we explore a simple collaboration between a LLM and a CRS for pre-sales dialogues, where one leads and the other assists.
% \begin{enumerate*}[label=(\roman*)]
    % \item {LLM leads and CRS assists.} We construct instructions for LLM based on the prediction results of CRS.
    % \item {CRS leads and LLM assists.} We design a module to encode the response of LLM and then add them to the CRS prediction probabilities.
% \end{enumerate*}
Specifically, we explore the effectiveness of collaborations on a real-world dataset of E-commerce pre-sales dialogues, namely U-NEED~\cite{liu2023u}.
U-NEED contains pre-sales dialogues in five top categories and supports four key tasks in E-commerce pre-sales dialogue:
% U-NEED contains four challenging tasks of E-commerce pre-sales dialogue:
% However, the complexities of real-life scenarios make it challenging to achieve a high quality pre-sales dialogue.
% User needs are difficult to be defined, and there are many expressions of user needs.
% Specifically, we design instructions and fine-tune LLMs to analyze their performance on four E-commerce pre-sales dialogue tasks: 
\begin{enumerate*}[label=(\roman*)]
    \item pre-sales dialogue understanding
    \item user needs elicitation
    \item user needs-based recommendation and
    \item pre-sales dialogue generation.
\end{enumerate*}
% We select ChatGLM-6B and Chinese-Alpaca-7B for the large model.
% For CRS we considered Bart-based and CPT-based models.
We select two popular open source LLMs, ChatGLM-6B and Chinese-Alpaca-7B, as well as two latest CRSs, Bart-based CRS and CPT-based CRS. We report experimental results for each combination of collaborations on the four challenging tasks.
% Specifically, we first train a CRS and fine-tune a LLM individually.
% Then, we continue training to let the CRS and the LLM learn to collaborate.
% Experimental results indicate that the effectiveness of collaborations of between a LLM and a CRS.
Experimental results demonstrate that the collaboration between LLM and CRS is effective on three tasks: pre-sales dialogue understanding, user needs elicitation and user needs-based recommendation.
Main contributions of this paper are as follows:
\begin{itemize}
    % \item We show the performance of LLMs in a real-world E-commerce scenario, i.e. pre-sales dialogue.  
    % \item To the best of our knowledge, we are the first to explore collaborations between LLMs and CRSs in a real-world scenario, namely E-commerce pre-sales dialogue.
    \item To the best of our knowledge, we are the first to explore collaboration between LLM and CRS in a real-world scenario, namely E-commerce pre-sales dialogue.
    \item We propose methods for two types of collaborations between LLMs and CRSs, i.e., \emph{CRS assisting  LLM} and \emph{LLM assisting  CRS}. 
    \item Extensive experimental results on a real-world E-commerce pre-sales dialogue dataset indicate the effectiveness and potential of collaborations between LLMs and CRSs.
    % , which can provide insights for the future research of leveraging LLMs in E-commerce customer service.
    % \item We explore a simple and effective collaboration between a LLM and a CRS for E-commerce pre-sales dialogues. 
    % for different tasks.
    % \item We propose a novel CRS that making recommendations given content generated by LLMs.
    % \item Our proposed collaborative approach achieves the best results on all tasks by combining the advantages of both LLMs and CRSs.
\end{itemize}

\section{Related Work}

We review the related work along two lines:
\begin{enumerate*}[label=(\roman*)]
    \item conversational recommendation and 
    \item \acfp{LLM} for recommendation.
\end{enumerate*}

% We review the related work along two lines: (i) conversational recommendation and (ii) large language models (LLMs) for recommendation.

\begin{figure*}[!t]
    \centering
    \includegraphics[width=1\textwidth]{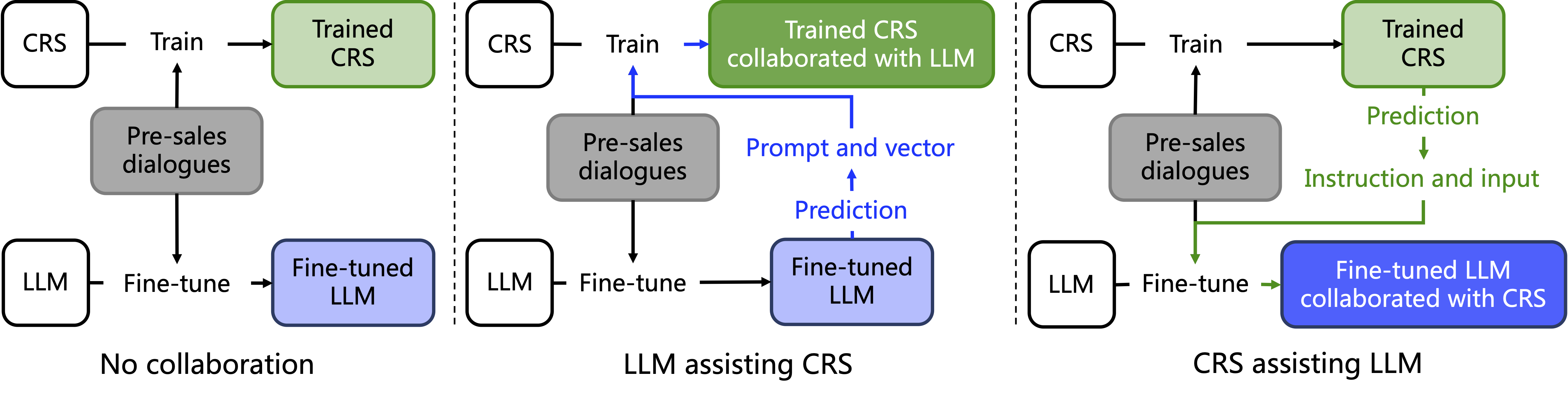}
    \caption{A comparison of the three types of collaboration between a \ac{CRS} and a \ac{LLM}. We explore the collaboration between \ac{LLM} and \ac{CRS}, i.e., \emph{\ac{LLM} assisting \ac{CRS}} and \emph{\ac{CRS} assisting \ac{LLM}}, and we compare the three in detail in \S\ref{sec:results}.}
    \label{fig:method}
\end{figure*}

\subsection{Conversational recommendation}
\Acfp{CRS} aim to provide real-time recommendations based on users' dynamic preferences through natural language interactions~\cite{gao-2021-advances,Jannach2021survey}.
Early work focus on:
\begin{enumerate*}[label=(\roman*)]
    \item question-based user preference elicitation~\cite{zou2020towards,hu2022learning}, 
    \item multi-turn conversational recommendation strategies~\cite{lei2020estimation, lei2020interactive},
    \item exploration-exploitation trade-offs~\cite{fu2021hoops, wong2021improving,zhang2020conversational},
    \item user preference modeling with external knowledge~\cite{zhou2022CCRS, chen-etal-2019-towards, ma-etal-2021-cr, zhou-etal-2021-crfr, ren2022variational}, 
    \item dialogue strategies~\cite{liu-etal-2020-towards-conversational, zhou2020towards, hayati2020inspired} and 
    \item generating persuasive responses~\cite{liang-etal-2021-learning-neural}.
\end{enumerate*}
Recently, some work~\cite{deng2023unified, wang-etal-2022-recindial, wang2022unified} utilize \acp{PLM} as the foundation to build unified CRSs, capable of performing various tasks using a single model, instead of multiple components. 

% Previous research has advanced CRSs, but 
The emergence of LLMs has undoubtedly impacted CRS-related researches. 
However, previous work barely explores the collaboration between conversational language models and CRSs on tasks related to conversational recommendation.
We investigate the collaboration between LLM and CRS in E-commerce pre-sales dialogues.

\subsection{LLMs for recommendation}

\Acfp{LLM}, such as GPT-3~\cite{gpt3}, InstructGPT~\cite{instructGPT}, PaLM~\cite{palm}, Bloom~\cite{bloom}, LLaMA~\cite{llama} and GLM~\cite{du2022glm}, have gained attention for their natural language understanding and generation capabilities~\cite{zhao2023survey}. 
Recent studies have examined the performance of ChatGPT~\cite{openai2022chatgpt} in tasks such as passage re-ranking~\cite{sun2023search} and recommendation~\cite{wang2023rethinking, liu2023chatgpt}. ChatGPT has also been applied to domains like augmenting recommender systems~\cite{gao2023chatrec}. Additionally, \citet{friedman2023leveraging} propose a roadmap for utilizing \ac{LLM} to build a controllable and explainable \ac{CRS} for YouTube videos.

Different from previous work using \ac{LLM} to enhance \ac{CRS}, we systematically investigate the effectiveness of combining \ac{LLM} and \ac{CRS}, i.e., \emph{\ac{LLM} assisting \ac{CRS}} and \emph{\ac{CRS} assisting \ac{LLM}}, which provides insights for future research on CRSs.

\section{Method}

\subsection{Overview}
In this paper, we explore the collaboration of a \acf{CRS} and a \acf{LLM}. 
Figure~\ref{fig:method} provides an illustration of our collaboration framework.

\header{LLM assisting CRS}
We leverage the prediction results of a \ac{LLM} to support a \ac{CRS}. 
Initially, we fine-tune a \ac{LLM} using pre-sales dialogues, following the method described in Section~\ref{sec:llm_for_tasks}. 
Subsequently, we incorporate the prediction results of the fine-tuned large model into the training process of the CRS, via prompts and vectors. 
For a detailed description, refer to Section~\ref{sec:llm_help_crs}.

\header{CRS assisting LLM}
We utilize the prediction results of a \ac{CRS} to assist a \ac{LLM}. 
Initially, we train a \ac{CRS} using pre-sales dialogues, following the approach outlined in Section~\ref{sec:crs_for_tasks}. 
Subsequently, we integrate the prediction results of the trained \ac{CRS} into the instructions and inputs to optimize the process of fine-tuning the LLM. For further details, see Section~\ref{sec:crs_help_llm}.

In this paper, we explore the effectiveness of collaboration by analyzing the impact of collaboration between \ac{CRS} and \ac{LLM} on the performance of four tasks in E-commerce pre-sales dialogue~\cite{liu2023u}.
These tasks include:
\begin{enumerate*}[label=(\roman*)]
    \item pre-sales dialogue understanding
    \item user needs elicitation
    \item user needs-based recommendation and
    \item pre-sales dialogue generation.
\end{enumerate*}
Due to space constraints, we provide detailed definitions of these tasks in Appendix~\ref{sec:app_task_formulation}.

\subsection{\acp{LLM} for E-commerce pre-sales dialogue}
\label{sec:llm_for_tasks}
We introduce the method of fine-tuning a \acf{LLM} using pre-sales dialogues.

\begin{figure*}[!t]
    \centering
    \includegraphics[width=1\textwidth]{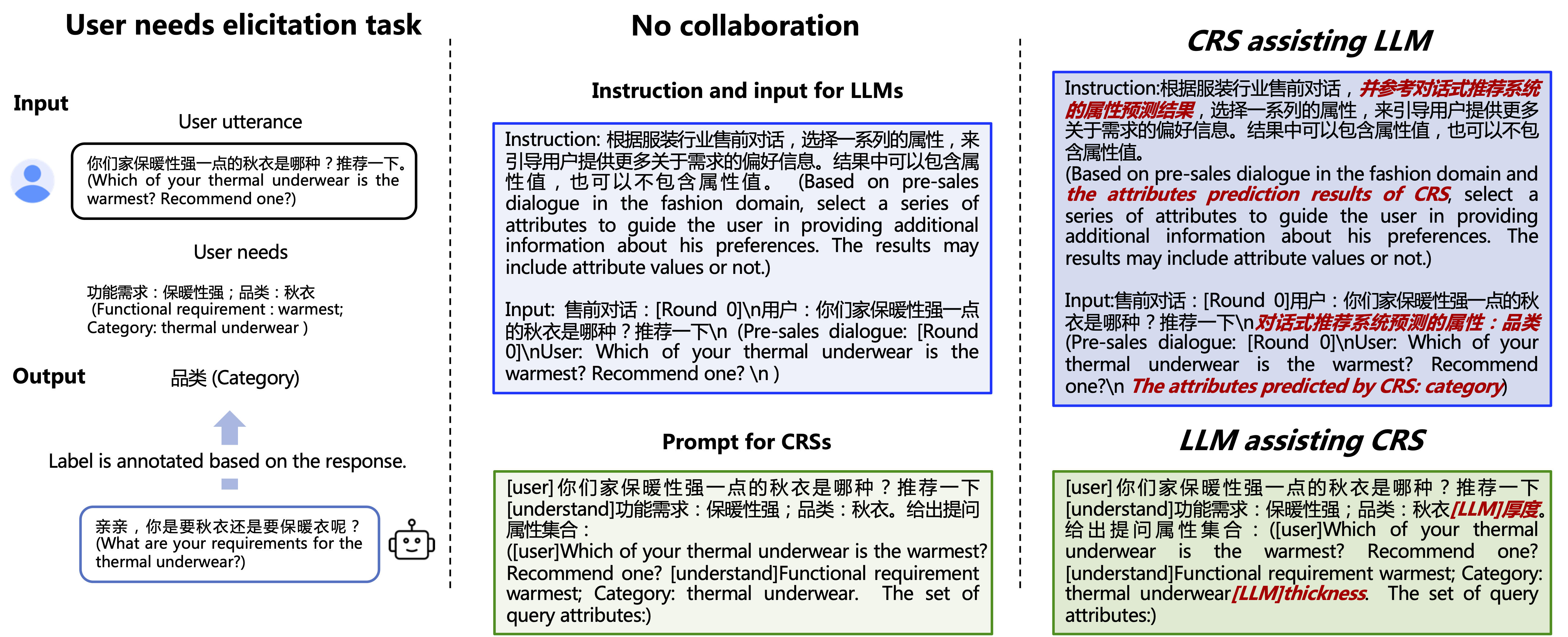}
    \caption{
        An example of collaboration between CRS and LLM on the user needs elicitation task. Left side shows the input and output of the task. 
        The middle displays data used to fine-tune a \ac{LLM} and train a \ac{CRS} independently. 
        The right side shows two cases of combining the two. 
        Collaboration content is highlighted in red italics.
        }
    \label{fig:sft_data}
\end{figure*}

\header{Instruction data}
Each sample within the training, validation, and test sets consists of ``instruction'', ``input'' and ``output''.
The ``instruction'' comprises several sentences that introduce the task's objective. 
The ``input'' contains the necessary information for completing the task. For instance, in the case of a user needs-based recommendation task, the ``input'' encompasses the user's needs, candidate products, and related product knowledge. 
The ``output'' remains consistent with the original task. 
Figure~\ref{fig:sft_data} shows an example of the instruction data corresponding to the user needs elicitation task. 
Additional examples of various tasks can be found in Appendix~\ref{sec:app_sft_examples}.
Note that the original user needs-based recommendation task involves numerous candidates, with each product possessing extensive attribute knowledge, the ``input'' surpasses the maximum input length permitted by \acp{LLM}. Consequently, in practice, we limit the number of candidates to 20.

\header{Base \acp{LLM} and fine-tuning}
We select ChatGLM and Chinese-Alpaca-7B as base \acp{LLM} due to their openness and commendable performance in Chinese basic semantic understanding.
ChatGLM is an open bilingual language model built upon General Language Model (GLM) framework~\cite{zeng2022glm}, with 6.2 billion parameters.\footnote{\url{https://github.com/THUDM/ChatGLM}}
LLaMA~\cite{llama} is a decoder-only, foundational large language model based on the transformer architecture~\cite{vaswani2017attention}. 
The Chinese LLaMA model is an extension of the original LLaMA model, incorporating an expanded Chinese vocabulary and undergoing secondary pre-training using Chinese data~\cite{chineseaplaca}.
We adopt the Chinese Alpaca model, which builds upon the aforementioned Chinese LLaMA model by incorporating instruction data for fine-tuning.\footnote{\url{https://github.com/ymcui/Chinese-LLaMA-Alpaca}} 
To carry out the fine-tuning process, we follow the official method provided by ChatGLM6B/Chinese-Alpaca-Plus-7B, using LoRA~\cite{lora}.

% The paper's writing and the content of the corresponding figures need revision to make them easier to understand. For example,

% Section 3.3 does not introduce each task in multi-task training, and there is no specific explanation or example for the design of prompts. Figure 3 and the Figures in the Appendix are also difficult to understand.
% The variable name definitions in Section 3.3 are problematic. 
%  is defined but not used. 
%  and 
%  are not defined. The user needs-based recommendation task corresponding to Eq. 2 and Eq. 3 is not defined. Section 3.5 has similar issues.
% In the experimental analysis section, the author uses tasks 1, 2, 3, and 4 to represent the tasks. Using task in Table 1, 2, 3, 4 or directly using the task name are more suitable.

\subsection{\acp{CRS} for E-commerce pre-sales dialogue}
\label{sec:crs_for_tasks}
We introduce the method of train a \acf{CRS} for pre-sales dialogues.
We adopt UniMIND~\cite{deng2023unified} as our base CRS, as it focus on multiple tasks in conversational recommendation.
% Our base CRS is UniMIND (Deng et al., 2023), which focuses on multiple tasks in conversational recommendation.
The recommendation candidates for UniMIND are movies. 
The movie title can be generated based on the prompt. 
However, in E-commerce pre-sales dialogue, the recommendation candidate is the product ID, which is difficult to be decoded directly. 
Therefore, for the user needs-based recommendation task we follow a traditional user representation-based approach~\cite{Kang2018sasrec}.
% In this paper, we modify the original UniMIND to address the four tasks of the E-commerce pre-sales dialogue.
% Concretely, there are three differences: (1) prompts for E-commerce pre-sales dialogue, (2) loss functions and (3) training process.

\header{Prompts}
Following~\citet{deng2023unified}, we define the inputs and outputs of the four tasks using a unified sequence-to-sequence paradigm.
%  we define the inputs and outputs of the four tasks as a unified sequence-to-sequence paradigm.
We use five special tokens to indicate information segments:
% We define five special tokens to indicate information segment.
\begin{enumerate*}[label=(\roman*)]
    \item \emph{[user]} indicates the utterance of the user.
    \item \emph{[system]} indicates the response from customer service staff.
    \item \emph{[understand]} indicates the needs contained in the user utterance, i.e., attributes and attribute values.
    \item \emph{[elicit]} indicates the attributes that the customer service staff plans to ask about user preferences.
    \item \emph{[recommend]} indicates the items that have been recommended by customer service staff.
\end{enumerate*}
For instance, the original input $X$ can be represented as follows:
\begin{equation}
    \begin{split}
        X_U &= \text{\small [user] }\,u_1\, \text{\small [understand] }\,d_1\, \text{\small [system] }\,s_1\,  \text{\small [understand]  } \\ & \,d_2\, \dots \text{\small [user] }\,u_i\,  \nonumber \\
        X_S &=\text{\small [user] }\,u_1\, \text{\small [understand] }\,d_1\, \text{\small [system] }\,s_1\,  \text{\small [understand]  } \\ &\,d_2\, \dots \text{\small [system] }\,s_i\,  \\
        X_A &=\text{\small [user] }\,u_1\, \text{\small [understand] }\,d_1\, {\text{\small [elicit] }\,a_1\,}  \text{\small [system] } \,s_1\, \\ & \text{\small [understand] }\,d_2\, \dots\text{\small [user] }\,u_i\,   \\
        X_R &=\text{\small [user] }\,u_1\, \text{\small [understand] }\,d_1\, {\text{\small [recommend] }\,e_1\,} \\ &  \text{\small [system] } \,s_1\,  \text{\small [understand] }\,d_2\, \dots\text{\small [user] }\,u_i\,   \\
        X_G &=\text{\small [user] }\,u_1\, \text{\small [understand] }\,d_1\, \text{\small [system] }\,s_1\,  \text{\small [understand]  } \\ &\,d_2\, \dots \text{\small [user] }\,u_i\, {\text{\small [elicit] }\,a_1\, }{\text{\small[recommend]}\,e_1\, } \\
    \end{split}
\end{equation}
where $\,u_i\,$ is the $i$-th utterance of the user, $\,s_i\,$ is the $i$-th response of customer service staff, $d_i$ is the $i$-th user needs, $a_i$ is the $i$-th attribute to be asked, and $e_i$is the $i$-th recommended product.
We adopt natural language prompt~\cite{raffel2020exploring} to indicate each task:
\begin{equation}
    \begin{split}
        Z_U &= \text{``Identify attributes and values:''} \nonumber  \\
        Z_A &= \text{``Select an attribute to ask:''} \\
        % Z_R &= \text{``Recommend an item:''} \\
        Z_G &= \text{``Generate a response:''} 
    \end{split}
\end{equation}

\header{Loss functions}
% \subsection{Collaboration between \ac{LLM} and \ac{CRS}}
% \label{sec:collaboration}
Following UniMIND~\cite{deng2023unified}, we design a unified \ac{CRS} with prompts learning and multitask learning for pre-sales dialogues.
\begin{eqnarray}
    \mathcal{L}_{\theta} = \mathbb{E}_{(X, Y, Z) \sim \mathcal{D}} \sum_{l=1}^L \log p_{\theta(y_l|y_{<l}, X, Z)},
\end{eqnarray}
where $\mathcal{D} = \{\mathcal{D}_U, \mathcal{D}_A, \mathcal{D}_G \}$ denote the train-set for three tasks: pre-sales dialogue understanding, user needs elicitation and pre-sales dialogue generation.
And $L$ is the length of the generated sequence.
For user needs-based recommendation task, the recommendation probability and loss function are defined as follows:
% Following~\cite{deng2023unified}, we also apply linear layer for mapping the representation from BART~\cite{lewis2020bart} encoder to candidate product space for user needs-based recommendation task:
\begin{eqnarray}
    r_i &=& \textbf{CLS}(\bm{e_i},\textbf{Enc}(X_R)) \\
    \mathcal{L}_R &=& - \sum_{i=1}^{|E|} e_i \log r_{i},
\end{eqnarray}
where $\bm{e_i}$ is the trainable item embedding, 
$E$ is collection of candidate products and $r_i$ is recommendation probability.
$\textbf{CLS}$(·) is a classifier, we apply linear layer in practice.
And $\textbf{Enc}$(·) is the encoder, we adopt two versions: BART~\cite{lewis2020bart} and CPT~\cite{shao2021cpt}.

\header{Training process}
We train a \ac{CRS} in two stages. 
In the first stage we train a \ac{CRS} only on the user needs-based recommendation task, i.e., $\mathcal{L} = \mathcal{L}_R$. 
Since BART and CPT do not have pre-sales conversation knowledge, this step aims to warm up CRS. 
In the second stage we continue to train a \ac{CRS} on all four tasks, i.e., $\mathcal{L} = \mathcal{L}_R + \mathcal{L}_{\theta}$.
% Moreover, we also consider UniMIND with CPT~\cite{shao2021cpt} model as a backbone.

\subsection{Collaboration 1: \ac{CRS} assisting \ac{LLM}}
\label{sec:crs_help_llm}
We introduce the method of CRS assisting LLM.
% Specifically, we first train a CRS model using the method in Section~\ref{sec:crs_for_tasks}.
% Next, we augment the original instructions and inputs of a LLM with the prediction results of the CRS model.
% Finally, we fine-tune a \ac{LLM} as described in \S\ref{sec:llm_for_tasks} using augmented instructions and inputs.
Initially, we train a CRS model following the approach outlined in Section ~\ref{sec:crs_for_tasks}. 
Subsequently, we enrich the original instructions and inputs of a LLM by incorporating the prediction outcomes from the CRS model. 
Finally, we fine-tune the LLM, as explained in \S~\ref{sec:crs_for_tasks}, utilizing the augmented instructions and inputs.

\header{Enhanced instruction and input}
% We transform the results of the trained \ac{CRS} into words and add them to the input of LLM. 
We convert the output of a trained CRS into text and incorporate it into the input of LLM.
% And we modify the instructions of \ac{LLM} to make it consider the voice form the trained \ac{CRS}.
We add additional instruction, i.e., completing the task requires considering the results of the CRS.
% In addition we inform the LLM to consider the results of CRS in the instruction.
An example is shown in the upper right corner of Figure~\ref{fig:sft_data}.
Note that in the user needs-based recommendation task, CRS can output a list of recommendations along with corresponding recommendation scores.
We rank the candidates according to their scores and include the ranking to the input of LLM.
For other tasks, we only consider the final output of the trained \ac{CRS}.
% there are multiple candidates, and we add the complete sequence of predictions of the trained \ac{CRS} to the input of \ac{LLM} in order to provide more information.

\subsection{Collaboration 2: \ac{LLM} assisting \ac{CRS}}
\label{sec:llm_help_crs}
We introduce the method of LLM assisting CRS.
% Specifically, we first fine-tune a LLM model using the method in Section~\ref{sec:llm_for_tasks}.
% Next, we use the prediction results from LLM to enhance the prompt and user representation.
% Finally, we train a \ac{CRS} as described in \S\ref{sec:crs_for_tasks}.
Initially, we fine-tune a LLM model using the technique described in Section~\ref{sec:llm_for_tasks}. Subsequently, we employ the prediction outcomes from LLM to enhance the prompt and user representation in the CRS model. Finally, we train a CRS model as outlined in \S~\ref{sec:crs_for_tasks}.

\header{Enhanced prompts}
We use the prediction results of \ac{LLM} to enhance the prompts of \ac{CRS}.
% For the output of \ac{LLM}, we encode it as a vector and add it to the task.
\begin{equation}
    \begin{split}
        X_U^{\prime} &= \text{\small [user] }\,u_1\, \text{\small [understand] }\,d_1\, \text{\small [system] }\,s_1\,  \text{\small [understand]  } \\ & \,d_2\, \dots \text{\small [user] }\,u_i\, \highlight{\text{\small [LLM] } \hat{a_i}, \hat{v_i}} \nonumber \\
        X_S^{\prime} &=\text{\small [user] }\,u_1\, \text{\small [understand] }\,d_1\, \text{\small [system] }\,s_1\,  \text{\small [understand]  } \\ &\,d_2\, \dots \text{\small [system] }\,s_i\, \highlight{\text{\small [LLM] } \hat{a_i}, \hat{v_i}} \\
        X_A^{\prime} &=\text{\small [user] }\,u_1\, \text{\small [understand] }\,d_1\, {\text{\small [elicit] }\,a_1\,}  \text{\small [system] } \,s_1\, \\ & \text{\small [understand] }\,d_2\, \dots\text{\small [user] }\,u_i\, \highlight{\text{\small [LLM] } \hat{a_i}}  \\
        X_G^{\prime} &=\text{\small [user] }\,u_1\, \text{\small [understand] }\,d_1\, \text{\small [system] }\,s_1\,  \text{\small [understand]  } \\ &\,d_2\, \dots \text{\small [user] }\,u_i\, {\text{\small [elicit] }\,a_1\, }{\text{\small[recommend]}\,e_1\, } \highlight{\text{\small [LLM] }\hat{\,s_i\,}} \\
    \end{split}
\end{equation}
For $X_U^{\prime}$ and $X_S^{\prime}$, $\hat{a_i}$ and $\hat{v_i}$ are attributes and values involved in utterance identified by \ac{LLM} for the $i$-th turn.
For $X_A^{\prime}$, $\hat{a_i}$ is the attribute in user needs elicitation for the $i$-th turn.
For $X_G^{\prime}$, $\hat{s_i}$ is the response for the pre-sales dialogue generation task generated by \ac{LLM}.
%  $\hat{s_i}$
%  is the output of \ac{LLM} for the $i$-th turn,  is the output of \ac{LLM} for the $i$-th turn in the user needs-based recommendation task.

\header{Enhanced representation} 
For recommendation task, we consider the embedding of the product recommended by the fine-tuned \ac{LLM}:
\begin{eqnarray}
    r_i^{\prime} &=& \textbf{CLS}(\bm{e_i},\textbf{Enc}(X_R), \highlight{\hat{\bm{e_i}}}) \\
    \mathcal{L}_R^{\prime} &=& - \sum_{i=1}^{|E|} e_i \log r_{i}^{\prime},
\end{eqnarray}
where $\hat{\bm{e_i}}$ is the embedding of the product recommended by the fine-tuned \ac{LLM}.

\section{Experimental Settings}

\subsection{Research questions}
To guide the remaining part of this paper, we set up two research questions:
% \begin{enumerate*}[label=(\roman*)]
%     % \item How reliably do the current generation of \acp{LLM} perform on pre-sales dialogue tasks?
%     \item Are LLM and CRS complementary? Does the combination of LLM and CRS improve performance?
%     \item How does the combination of CRS and LLM perform on different tasks and different categories? What are the differences between different collaboration methods?
% \end{enumerate*}
\begin{itemize}
    \item Are LLM and CRS complementary? Does the combination of LLM and CRS improve performance?
    \item How does the combination of CRS and LLM perform on different tasks and different categories? What are the differences between different collaboration methods?
\end{itemize}
% In the Results and Analysis section, we analyze experimental results of each task in turn and obtain findings based on these two questions.
In the Results and Analysis section, we systematically examine the outcomes of each task to answer the aforementioned two research questions.

\subsection{Dataset}

We conduct experiments on U-NEED~\cite{liu2023u}.
% \acfi{U-NEED} from real-world E-commerce scenarios. 
% \ac{U-NEED} consists of 3 types of resources:
% \begin{enumerate*}[label=(\roman*)]
%     \item 7,698 fine-grained annotated pre-sales dialogues in 5 top categories
%     \item 333,879 user behaviors and 
%     \item 332,148 product knowledge tuples.
% \end{enumerate*}
% To facilitate the research of \ac{UNECR}, we propose 5 critical tasks:
% \begin{enumerate*}[label=(\roman*)]
%     \item pre-sales dialogue understanding
%     \item user needs elicitation 
%     \item user needs-based recommendation
%     \item pre-sales dialogue generation and 
%     \item pre-sales dialogue evaluation.
% \end{enumerate*}
U-NEED consists of 7,698 fine-grained annotated pre-sales dialogues, which consist of 1662, 1513, 1135, 1748, and 1640 dialogues in \textit{Beauty}, \textit{Phones}, \textit{Fashion}, \textit{Shoes} and \textit{Electronics} categories respectively.
% We follow the data split proposed in U-NEED.
We follow the partition of the training set, validation set and test set proposed in U-NEED.
% Following~\cite{liu2023u}, 
% \ac{U-NEED} spans 30 days, and we utilize dialogues of the first 24 days as the training set, dialogues of the last 3 days as the test set, and the rest as the validation set.

\begin{table*}[!t]
    \caption{Performance of baseline methods on pre-sales dialogue understanding task in 3 typical categories: \textit{Beauty}, \textit{Fashion} and \textit{Shoes}. 
    Baseline results marked with * are taken from U-NEED~\cite{liu2023u}.
    CLLM is short for ChatGLM and ALLM is short for Chinese-Alpaca.
    BCRS is short for UniMIND(BART) and CCRS is short for UniMIND(CPT).
    The best results are highlighted in bold.
    }
    \resizebox{\textwidth}{!}{
    \begin{tabular}{lcccccccccccc}
    \toprule
    & \multicolumn{3}{c}{Beauty} & \multicolumn{3}{c}{Shoes} & \multicolumn{3}{c}{Phones} & \multicolumn{3}{c}{All 5 categories}  \\ \cmidrule(r){2-4}\cmidrule(r){5-7}\cmidrule(r){8-10} \cmidrule{11-13}
    Methods & P & R & F1 & P & R & F1 & P & R & F1 & P & R & F1 \\ \midrule
    Bert* & 0.5355 & 0.6284&0.5782&0.5851&0.7020&0.6382 & 0.4212&0.5384&0.4726&0.4549&0.5652&0.5041  \\
    Bert+CRF* & 0.6731&0.6802&0.6766&0.7302&0.7703&0.7497&0.5620&0.5923&0.5768&0.6688&0.6530&0.6608   \\
    Bert+BiLSTM+CRF* &  0.7282&0.7481&0.7380&0.7870&0.8101& 0.7984&0.6701&0.6990& 0.6843&0.6892& 0.6875 & 0.6884 \\
    \midrule
    \textbf{\emph{No collaboration}} \\
    UniMIND(BART) &0.6443&0.6000&0.6085&0.7711&0.7417&0.7483&0.7522&0.7406&0.7407&0.7188&0.6933&0.6978\\
    UniMIND(CPT)& 0.5994&0.5420&0.5565&0.7468&0.6836&0.6889&0.7110&0.6907&0.6959&0.6807&0.6451&0.6539 \\ 
    % \midrule
    ChatGLM & 0.7858 & 0.7797 & 0.7777 & 0.8265 & 0.8307 & 0.8248 & 0.7805 & 0.7792 & 0.7760 & 0.7968 & 0.7936 & 0.7910 \\   
    Chinese-Alpaca & 0.7409 & 0.7310 & 0.7316 & 0.8032 & 0.7868 & 0.7899 & 0.7363 & 0.7178 & 0.7238 & 0.7568 & 0.7378 & 0.7425\\
    \midrule
    \textbf{\emph{LLM assisting CRS}} \\
    CLLM-BCRS &0.6502&0.5848&0.6004&0.7725&0.7171&0.7318&0.7504&0.7152&0.7246&0.7173&0.6665&0.6796\\
    CLLM-CCRS & 0.6101&0.5346&0.5545&0.7663&0.7218&0.7338&0.7183&0.6976&0.7018&0.7023&0.6539&0.6666\\
    ALLM-BCRS & 0.6255  & 0.5688  & 0.5835  & 0.7746  & 0.7120  & 0.7302  & 0.7311  & 0.7108  & 0.7159  & 0.7088  & 0.6653  & 0.6765 \\
    ALLM-CCRS & 0.5730  & 0.5307  & 0.5410  & 0.7048  & 0.6658  & 0.6768  & 0.6833  & 0.6635  & 0.6691  & 0.6629  & 0.6277  & 0.6369 \\
    \midrule
    \textbf{\emph{CRS assisting LLM}} \\
    BCRS-CLLM & 0.7900 &0.7879 &0.7824 &\textbf{0.8521} &\textbf{0.8511} &\textbf{0.8473} &\textbf{0.8222} &\textbf{0.8220} &\textbf{0.8179} &\textbf{0.8105} &\textbf{0.8065} &\textbf{0.8033}   \\
    CCRS-CLLM &\textbf{0.7940} &\textbf{0.7926} &\textbf{0.7878} &0.8372 &0.8378 &0.8326 &0.7911 &0.7866 &0.7847 &0.7927 &0.7897 &0.7864   \\
    BCRS-ALLM & 0.7772 &0.7462 &0.7542 &0.8062 &0.7690 &0.7770 &0.7662 &0.7160 &0.7311 &0.7700 &0.7310 &0.7414 \\
    CCRS-ALLM & 0.7600 &0.7272 &0.7354 &0.8036 &0.7621 &0.7738 &0.7392 &0.6978 &0.7086 &0.7569 &0.7168 &0.7276 \\
    \bottomrule
    \end{tabular}
    }
    \label{tab:experiment_task1}
\end{table*}

\subsection{Baseline methods}
For each task, baseline methods consist of typical methods, \ac{CRS} methods and \ac{LLM} methods.

\header{Typical methods}
We select typical methods for the four tasks following~\cite{liu2023u}.
Specifically, we select Bert, Bert+CRF, Bert+BiLSTM+CRF as baselines for pre-sales dialogue.
For user needs elicitation task, we select DiaMultiClass and DiaSeq as baselines.
For user needs-based recommendation task, we choose Bert, SASRec, TG-CRS.
And we select GPT-2 and KBRD as baseline methods for pre-sales dialogue generation task.
For the limited space, we put the description of each typical methods in the Appendix~\ref{sec:app_baselines}.
We select UniMIND(BART)~\cite{deng2023unified} and UniMIND(CPT) as CRS methods.
For LLM methods, we select ChatGLM~\cite{zeng2022glm} and Chinese-Alpaca~\cite{chineseaplaca}.
For combination of \ac{LLM} and \ac{CRS}, we define eight variants.
We put the description of each variant in the Appendix~\ref{sec:app_variants}.

\subsection{Evaluation metrics}

We adopt the evaluation metrics used in U-NEED~\cite{liu2023u}. 
Specifically, we select precision, recall and f1 score as evaluation metrics for pre-sales dialogue understanding and user needs elicitation.
% For user needs elicitation task, we select DiaMultiClass and DiaSeq as baselines.
For user needs-based recommendation task, we choose Hit@K and MRR@K.
And we adopt automatic and human evaluation for pre-sales dialogue generation task.
For automatic evaluation, we use Distinct-n. 
And for human evaluation, we measure the informativeness and relevance of generated response.
For the limited space, we put the description of each metric in the Appendix~\ref{sec:app_evaluation_metrics}.

\begin{table*}[!t]
    \caption{Performance of baseline methods on user needs elicitation task in 3 typical categories: \textit{Beauty}, \textit{Fashion} and \textit{Shoes}. 
    Baseline results marked with * are taken from U-NEED~\cite{liu2023u}.
    CLLM is short for ChatGLM and ALLM is short for Chinese-Alpaca.
    BCRS is short for UniMIND(BART) and CCRS is short for UniMIND(CPT).
    The best results are highlighted in bold.
    }
    \resizebox{\textwidth}{!}{
    \begin{tabular}{lcccccccccccc}
    \toprule
    & \multicolumn{3}{c}{Beauty} & \multicolumn{3}{c}{Shoes} & \multicolumn{3}{c}{Phones} & \multicolumn{3}{c}{All 5 categories}  \\ \cmidrule(r){2-4}\cmidrule(r){5-7}\cmidrule(r){8-10} \cmidrule{11-13}
    & P & R & F1 & P & R & F1 & P & R & F1 & P & R & F1 \\ 
    \midrule
    DiaMultiClass* & 0.4037& \textbf{0.7228}&\textbf{0.5054}&0.3361&\textbf{0.4131}&0.3423&0.4534&\textbf{0.5212}&0.4585&0.3222&\textbf{0.4966}&0.3662 \\
    DiaSeq* & 0.4761& 0.4272&0.4424&0.3992&0.3305&0.3498&0.4414&0.3789&0.3966&0.3555&0.2996&0.3153 \\
    \midrule
    \textbf{\emph{No collaboration}} \\
    UniMIND(BART) & 0.4979&0.4305&0.4518&0.4367&0.3827&0.3927&0.5513&0.4973&0.5061&0.4301&0.3881&0.3943 \\
    UniMIND(CPT)& 0.4022&0.3575&0.3657&0.4388&0.3774&0.3906&0.4946&0.4432&0.4515&0.4021&0.3575&0.3657\\ 
    % \midrule
    ChatGLM &  0.4348  & 0.4057  & 0.4055  & 0.4054  & 0.3496  & 0.3621  & 0.4712  & 0.4541  & 0.4441  & 0.3683  & 0.3422  & 0.3380 \\   
    Chinese-Alpaca & 0.2723  & 0.2362  & 0.2475  & 0.4020  & 0.3440  & 0.3588  & 0.4946  & 0.4613  & 0.4604  & 0.3513  & 0.3132  & 0.3191 \\
    \midrule
    \textbf{\emph{LLM assisting CRS}} \\
    CLLM-BCRS & \textbf{0.5106}&0.4413&0.4617&0.4510&0.3837&\textbf{0.4016}&\textbf{0.5703}&0.4982&\textbf{0.5168}&\textbf{0.4518}&0.3963&\textbf{0.4095}\\
    CLLM-CCRS & 0.4128&0.3405&0.3583&\textbf{0.4531}&0.3735&0.3955&0.5243&0.4541&0.4737&0.4056&0.3405&0.3583\\
    ALLM-BCRS & 0.4702  & 0.4149  & 0.4289  & 0.4490  & 0.3827  & 0.3969  & 0.5297  & 0.4865  & 0.4863  & 0.4258  & 0.3866  & 0.3903  \\
    ALLM-CCRS & 0.4043  & 0.3362  & 0.3574  & 0.4490  & 0.3661  & 0.3884  & 0.5505  & 0.4928  & 0.5009  & 0.4138  & 0.3519  & 0.3671 \\
    \midrule
    \textbf{\emph{CRS assisting LLM}} \\
    BCRS-CLLM & 0.4908 &0.4319 &0.4450 &0.3918 &0.3280 &0.3429 &0.4923 &0.4788 &0.4645 &0.4190 &0.3723 &0.3796  \\
    CCRS-CLLM &0.4092 &0.3766 &0.3783 &0.4082 &0.3439 &0.3590 &0.4495 &0.4356 &0.4211 &0.4020 &0.3710 &0.3704 \\
    BCRS-ALLM & 0.2660 &0.2383 &0.2428 &0.2490 &0.1914 &0.2067 &0.2568 &0.2432 &0.2378 &0.2350 &0.1904 &0.1994  \\
    CCRS-ALLM & 0.1681 &0.1489 &0.1539 &0.2388 &0.1826 &0.1951 &0.2027 &0.1874 &0.1823 &0.2071 &0.1632 &0.1730  \\
    \bottomrule
    \end{tabular}
    }
    \label{tab:experiment_task2}
\end{table*}
\section{Results and Analysis}
\label{sec:results}
We conduct extensive experiments to explore the performance of collaborations of \acp{LLM} and \acp{CRS} on four tasks.
% of E-commerce pre-sales dialogue.
We analyze impacts of collaborations on each task in turn. 

\subsection{Impacts of collaborations on pre-sales dialogue understanding}
Based on Table~\ref{tab:experiment_task1} we have the following observations:
\begin{enumerate*}[label=(\roman*)]
    \item \pemph{\acp{LLM} perform well in understanding user needs.} ChatGLM substantially outperforms classical baseline methods and \acp{CRS} on all metrics in all categories. 
    The second best method is Chinese-Alpaca, which outperforms the strongest baseline method, Bert+BiLSTM+CRF, on most metrics. We attribute this to the strong capability of the \acp{LLM} for dialogue understanding. 
    \item \pemph{In collaborations with \acp{CRS}, \acp{LLM} perform even better in understanding user needs.}
    BCRS-CLLM outperforms ChatGLM in all category and all metrics, especially in \textit{Shoes} and \textit{Phones}.
    Similarly we observe that BCRS-ALLM outperforms Chinese-Alpaca in some metrics.
    By carefully comparing the results of the four combinations of ``CRS assisting LLM'' with the results of the four methods of ``No collaboration'', we find that a better performing CRS improves the performance, while a worse performing CRS degrades the performance. 
    Since \acp{LLM} are usually very sensitive to inputs, we think that the former may provide useful reference information to complement what the \acp{LLM} do not take into account. The latter, on the other hand, may bring in noises that disturb the judgments of the \acp{LLM}.
    % Moreover, the performance of \acp{LLM} is further improved by considering the prediction results of \acp{CRS}. 
    % Note that traditional CRS methods struggle to model the fine-grained needs and preferences in a dialogue. 
    % The emergence of \acp{LLM}, on the other hand, makes it possible for a pre-sales shopping robot to truly understand users' needs and preferences. 
    % In practice,  "CRS assisting LLM" approach may be a viable solution based on the results in Table 1.
    \item \pemph{The improvement in understanding user needs brought by the collaboration of CRS to \acp{LLM} varies across categories.}
    In the \textit{Shoes} and \textit{Phones} categories, BCRS-CLLM significantly outperforms ChatGLM with the collaborations of \acp{CRS}.
    While in the \textit{Beauty} category, BCRS-CLLM has only a minor improvement compared to ChatGLM. 
    We think this may be due to the fact that user needs in the \textit{Beauty} category are usually focused on specific attributes such as ``skin type''.
\end{enumerate*}

\begin{table*}[!t]
    \caption{Performance of baseline methods on user needs-based recommendation task in 3 typical categories: \textit{Beauty}, \textit{Fashion} and \textit{Shoes}. 
    H@K and M@K refer to Hit@K and MRR@K.
    Acc. refers to accuracy.
    CLLM is short for ChatGLM and ALLM is short for Chinese-Alpaca.
    BCRS is short for UniMIND(BART) and CCRS is short for UniMIND(CPT).
    Since LLM methods only recommend 1 product, H@5 and M@5 cannot be calculated.
    The best results are highlighted in bold.
    }
    \resizebox{\textwidth}{!}{
    \begin{tabular}{lcccccccccccc}
    \toprule
    & \multicolumn{3}{c}{Beauty} & \multicolumn{3}{c}{Shoes} & \multicolumn{3}{c}{Phones} & \multicolumn{3}{c}{All 5 categories}  \\ \cmidrule(r){2-4}\cmidrule(r){5-7}\cmidrule(r){8-10} \cmidrule{11-13}
    & Acc. & H@5 & M@5 & Acc. & H@5 & M@5 & Acc. & H@5 & M@5 & Acc. & H@5 & M@5 \\ \midrule
    Bert& 0.0123 & 0.0185 & 0.0141 & 0.0046 & 0.0138 & 0.0072 & 0.0326 & 0.0688 & 0.0463 & 0.0118 & 0.0215 & 0.0151\\
    SASRec& 0.1108 & 0.2831 & 0.1711 & 0.0399 & 0.1121 & 0.0668 & 0.0761 & 0.2681 & 0.1475 & 0.0976 & 0.2532 & 0.1556  \\ 
    TG-CRS& 0.1323 & 0.3354 & 0.2034 & 0.1275 & 0.2396 & 0.1692 & 0.2564 & 0.4928 & 0.3347 & 0.1744 & 0.3074 & 0.2244 \\
    \midrule
    \textbf{\emph{No collaboration}} \\
    UniMIND(BART) &0.2154&0.6246&0.3654&0.2458&0.5315&0.3483&0.3478&0.6449&0.4608&0.2398&0.5440&0.3510\\
    UniMIND(CPT)  &0.2554&0.6246&\textbf{0.3915}&0.2826&0.5438&0.3779&0.3043&0.6594&0.4460&0.2639&0.5617&0.3737\\ 
    % \midrule
    ChatGLM & 0.2123  & - & - & 0.3810  & - & - & 0.0580  & - & - & 0.2226  & - & - \\   
    Chinese-Alpaca & 0.1415  & - & - & \textbf{0.3917}     & - & -  & 0.0036 & - & - &0.2157     & - & -
    \\
    \midrule
    \textbf{\emph{LLM assisting CRS}} \\
    CLLM-BCRS & 0.2462&0.6062&0.3741&0.2565&0.5376&0.3594&0.3804&0.7536&0.5204&0.2623&0.5617&0.3694\\
    CLLM-CCRS & \textbf{0.2585}&0.6308&0.3913&0.2657&0.5438&0.3690&0.4058&0.7138&0.5319&0.2768&0.5665&0.3835\\
    ALLM-BCRS & 0.2554 &0.6092 &0.3822 &0.2550 &0.5223 &0.3554 &0.3587 &\textbf{0.7681} &0.5196 &0.2623 &0.5606 &0.3721 \\
    ALLM-CCRS & 0.2430 &\textbf{0.6369} &0.3747 &0.2750 &\textbf{0.5515} &\textbf{0.3788} &\textbf{0.4312} &0.7609 &\textbf{0.5959} &\textbf{0.2822} & \textbf{0.5832} & \textbf{0.3919} \\
    \midrule
    \textbf{\emph{CRS assisting LLM}} \\
    BCRS-CLLM & 0.1600 &-&-&0.2442 &-&-&0.3478 &-&-&0.2264 &-&- \\
    CCRS-CLLM & 0.1785 &-&-&0.2642 &-&-&0.3043 &-&-&0.2479 &-&- \\
    BCRS-ALLM & 0.2000 &-&-&0.2458 &-&-&0.3297 &-&-&0.2307 &-&- \\
    CCRS-ALLM & 0.2369 &-&-&0.2688 &-&-&0.2754 &-&-&0.2532 &-&- \\
    \bottomrule
    \end{tabular}
    }
    \label{tab:experiment_task3}
\end{table*}

\subsection{Impacts of collaborations on user needs elicitation}
Based on Table~\ref{tab:experiment_task2} we have the following observations:
\begin{enumerate*}[label=(\roman*)]
    % \item \pemph{\acp{CRS} and \acp{LLM} show limited performance in user needs elicitation.}
    \item \pemph{\acp{LLM} do not exhibit superb performance on user needs elicitation.} On the average results of the 5 categories, UniMIND(BART) achieves the best performance, followed by the classical method DialMultiClass and UniMIND(CPT). Moreover, DiaMultiClass beats all the methods on Recall metrics in Beauty, Shoes, and Mobile categories. This indicates that in E-commerce pre-sales dialogue, the performance of methods that make decisions with the help of generative models, e.g. BART and \acp{LLM}, is somewhat limited. DialMultiClass doesn't require a large number of parameters and doesn't need to be trained for a long period of time. Compared to making decisions with \acp{LLM}, DialMultiClass still has considerable strengths in real-world production environments.
    \item \pemph{Collaborations between ChatGLM and UniMIND (BART) can improve their respective performance in user needs elicitation.} Specifically, CLLM-BCRS outperforms UniMIND(BART) on all metrics for all categories. Moreover, CLLM-BCRS beats all methods on six metrics.
    Similarly, BCRS-CLLM outperforms ChatGLM in all categories except \textit{Shoes}. Specifically, BCRS-CLLM achieves a 12.3\% improvement over ChatGLM on the average F1 score across all 5 categories. 
    Based on this, we see that the output of ChatGLM are beneficial for UniMIND(BART) and vice versa.
    % chatglm and bcrs benefit each other
    % Chatglm improves the performance on recall. 
    \item \pemph{Chinese-Alpaca and ChatGLM exhibit major differences in their performance in collaborations with \acp{CRS}.} 
    Specifically, CCRS-ALLM and BCRS-ALLM achieve the worst and second-worst performance on almost all metrics in all categories. This implies that the predicted results of \acp{CRS} cause a large disruption to Chinese-Alpaca. The two combinations of \acp{CRS} assisting ChatGLM, i.e., BCRS-CLLM and CCRS-CLLM, however, perform well. 
    We think that the differences between ChatGLM and Chinese-Alpaca in collaborating with \acp{CRS} come from their base models and fine-tuning data.
% ChatGLM is based on GLM and adopts the technology similar to ChatGPT. Chinese-Alpaca is based on LLaMa and extended with Chinese data and dialogue data. 
% The capacity of \acp{LLM} to provide decision-making information in certain ways (e.g., chain of thoughts) is a key point of exploration in \acp{LLM}. In this work, we explore whether \acp{LLM} can select attributes that can elicit users' preferences. The results in Table 2 show the state-of-the-art performance of \acp{LLM} in assisting \acp{CRS} on the Precision metric. However, in the Recall metric it is the classical approach applied in industry that achieves state-of-the-art performance. Note that ChatGLM assisting \acp{CRS} improves performance, while Chinese-Alpaca assisting \acp{CRS} decreases performance. This may be an accumulation of errors due to the limited performance of Chinese-Alpaca on Task 2. The collaboration between \acp{LLM} and \acp{CRS} in terms of strategies leaves a lot of room for exploration. We plan to explore more sophisticated ways of collaboration in future work, such as \acp{LLM} and \acp{CRS} engaging in debates to decide which attributes should be chosen at the moment.
\end{enumerate*}

\subsection{Impacts of collaborations on user needs-based recommendation}
Based on Table~\ref{tab:experiment_task3} we have the following observations:
\begin{enumerate*}[label=(\roman*)]
    % \item \pemph{\acp{LLM} (i.e. generative recommendation) can boost \acp{CRS} to achieve the best performance.}
    % Note that although we fine-tuned the \acp{LLM} with training data, the \acp{LLM} providing product recommendations is not a classification task. \acp{LLM} select and recommend product from a given candidate list based on understanding the given instruction and input. \acp{CRS}, on the other hand, model the product recommendation as a classification task, and its prediction is computed based on the relations between user representation and candidate product representation. We consider these to be two different approaches to providing recommendations. The results in Table 3 show that modeling the relations between user representations and candidate item representations is still effective, and the two ways of providing recommendations are mutually helpful. In detail, \acp{LLM} can enhance \acp{CRS} to achieve state-of-the-art performance, while \acp{CRS} can also help \acp{LLM} to get moderate performance. We believe that this is inspiring to study generative recommender systems for E-commerce scenarios.
    \item \pemph{\acp{LLM} show the potential for user needs-based recommendation.} Specifically, \acp{LLM}, i.e., Chinese-Alpaca and ChatGLM, achieve the best and second best performance significantly outperforming all the methods on the Accuracy in Shoes category, respectively. They also achieve performance over classical methods on the results of all 5 categories. Based on this, we think that \acp{LLM} can somewhat provide suitable recommendations when the candidate range is small (the number of candidate products in Table~\ref{tab:experiment_task3} is 20).
    \item \pemph{With the collaboration of \acp{LLM}, the recommendation performance of \acp{CRS} can be improved.} 
    Specifically, on the average results across all 5 categories, ALLM-CCRS achieves 6.9\%, 3.8\%, and 4.9\% improvements on Accuracy, Hit@5, and MRR@5, respectively, compared to UniMIND (CPT). Similarly, on average results across all 5 categories, ALLM-BCRS achieves 9.4\%, 3.0\%, and 6.0\% improvements on Accuracy, Hit@5, and MRR@5, respectively, when compared to UniMIND(BART). Note that \acp{LLM} provide recommendations in a different way than \acp{CRS} do. \acp{LLM} provide recommendations relying on given inputs, i.e., a recommended product is semantically related to user needs in some way. \acp{CRS}, on the other hand, model a representation of both and learn the implicit relationship between the two to compute the probability of a product being recommended. The above improvements are only that \acp{CRS} consider the representations of the recommended products given by the \acp{LLM}. We believe that collaborations between \acp{LLM} and \acp{CRS} on recommendation tasks go far beyond this and are a direction worth exploring.
    \item \pemph{With collaborations with \acp{CRS}, \acp{LLM} can achieve comparable recommendation performance.} Specifically, in the \textit{Phones} category, ChatGLM and Chinese-Alpaca have very poor recommendation performance.
    In contrast, the four methods of ``CRS assisting LLM'' achieve the performance close to that of \acp{CRS}.
    % in the \textit{Phones} category. 
    Based on this, we think that when the recommendation performance of \acp{LLM} is very poor in a certain domain, a collaborative approach could make \acp{LLM} to achieve performance close to that of \acp{CRS}.
\end{enumerate*}

\begin{table*}[!t]
    \caption{Performance of baseline methods on pre-sales dialogue generation task in 3 typical categories: \textit{Beauty}, \textit{Fashion} and \textit{Shoes}. 
    CLLM is short for ChatGLM and ALLM is short for Chinese-Alpaca.
    BCRS is short for UniMIND(BART) and CCRS is short for UniMIND(CPT).
    Info. and Rel. refer to informativeness and relevance.
    The best results are highlighted in bold.
    }
    \resizebox{\textwidth}{!}{
    \begin{tabular}{lcccccccccccc}
    \toprule
    & \multicolumn{3}{c}{Beauty} & \multicolumn{3}{c}{Shoes} & \multicolumn{3}{c}{Phones} & \multicolumn{3}{c}{All 3 categories}  \\ \cmidrule(r){2-4}\cmidrule(r){5-7}\cmidrule(r){8-10} \cmidrule{11-13}
    & Dist-1 & Rel. & Info. & Dist-1 & Rel. & Info. & Dist-1 & Rel. & Info. & Dist-1 & Rel. & Info. \\
    \midrule
    GPT-2&  0.6195&1.8800&1.8400&0.6504&2.1933&2.1467&0.6305&1.9867&1.8067&0.6335&2.0200&1.9311  \\  
    KBRD& 0.5753&2.9933&2.5067&0.5639&3.3467&2.8800&0.6034&3.1000&2.7533&0.5809&3.1467&2.7133  \\ 
    \midrule
    \textbf{\emph{No collaboration}} \\
    UniMIND(BART) & 0.8611&3.7933&3.3600&0.8897&3.9133&3.6933&0.8299&3.6667&3.6200&0.8521&3.7911&3.5578  \\
    UniMIND(CPT)& 0.8578&3.7333&3.2600&0.9080&3.9333&3.7800&0.8472&\textbf{3.8000}&\textbf{3.7400}&0.8583&3.8222&3.5933  \\ 
    ChatGLM & 0.8169&3.7467&3.4533&0.8594&3.7000&3.4400&0.8411&3.6200&3.4533&0.8152&3.6889&3.4489  \\   
    Chinese-Alpaca & \textbf{0.9131}&3.6667&3.2467&\textbf{0.9124}&3.8733&3.6667&\textbf{0.9109}&3.6733&3.6133&\textbf{0.8927}&3.7378&3.5089  \\
    \midrule
    \textbf{\emph{LLM assisting CRS}} \\
    CLLM-BCRS & 0.8686&3.8467&3.4933&0.8986&3.9133&3.7600&0.8460&3.7267&3.6000&0.8611&3.8289&\textbf{3.6178}  \\
    CLLM-CCRS & 0.8633&3.7667&\textbf{3.5000}&0.8975&3.9267&3.7200&0.8568&3.6267&3.5267&0.8587&3.7734&3.5822 \\
    ALLM-BCRS & 0.8700&3.7667&3.4000&0.8959&3.9200&\textbf{3.7867}&0.8505&3.7333&3.6533&0.8650&3.8067&3.6133  \\
    ALLM-CCRS & 0.8792&3.8467&3.4467&0.9086&\textbf{3.9933}&\textbf{3.7867}&0.8640&3.6400&3.5933&0.8699&3.8267&3.6089  \\
    \midrule
    \textbf{\emph{CRS assisting LLM}} \\
    BCRS-CLLM & 0.8371&3.6867&3.3867&0.8735&3.9400&3.7667&0.8544&3.6733&3.5800&0.8365&3.7667&3.5778  \\
    CCRS-CLLM & 0.8324&3.5000&3.0867&0.8714&3.9200&3.7800&0.8355&3.5933&3.4533&0.8256&3.6711&3.4400  \\
    BCRS-ALLM & 0.8910&3.6933&3.2200&0.9011&3.7867&3.6133&0.9040&3.6067&3.5400&0.8804&3.6956&3.4578  \\
    CCRS-ALLM & 0.8898&\textbf{3.8600}&3.3400&0.8951&3.9333&3.6733&0.8988&3.7400&3.5667&0.8826&\textbf{3.8444}&3.5267   \\
    \bottomrule
    \end{tabular}
    }
    \label{tab:experiment_task4}
\end{table*}

\subsection{Impacts of collaborations on pre-sales dialogue generation}
Based on Table~\ref{tab:experiment_task4} we have the following observations:
\begin{enumerate*}[label=(\roman*)]
    \item \pemph{\acp{CRS} and \acp{LLM} show comparable performance in pre-sales dialogue generation.}
    In Shoes, Beauty, and Phones categories, Chinese-Alpaca achieves the best performance on Dist-1. This indicates that Chinese-Alpaca can generate more diverse responses.
    While in most cases, the responses generated by \acp{CRS} are more relevant and informative than those generated by \acp{LLM}.
    In addition, neither the \acp{LLM} nor the \acp{CRS} generate responses that beat the ground truth responses provided by customer service staff.
    \item \pemph{Collaborations between \acp{LLM} and \acp{CRS} show marginal effects on pre-sales dialogue generation. } Specifically, the methods of collaborations between \acp{LLM} and \acp{CRS}, i.e., ``LLM assisting CRS'' and ``CRS assisting LLM'', achieve the best performance on most of the metrics. However, the improvement from collaboration is marginal compared to \acp{CRS} or \acp{LLM}.
    We believe this may be due to the fact that \acp{LLM} and UniMIND are relatively close in their approaches to generating responses, i.e., both are based on pre-trained language models and prompts.
    Therefore, collaborations between two similar approaches does not have much impact. 
    In future work, we plan to consider \acp{CRS} that focus on generating persuasive reasons for recommendations, e.g., NTRDs that introduce words related to the recommended items in the decoding process. Intuitively, collaborations between such \acp{CRS} and \acp{LLM} may work out well.
    % \item It is widely recognized that the \acp{LLM} are much better at generating responses than the smaller models. However, Table~\ref{tab:experiment_task4} demonstrates that the performance of the two is comparable and does not exceed the human standard response. In the human evaluation: a score of 4 (or 5) means that a model-generated response meets (or exceeds) the ground truth response, respectively. And there is no major improvement in collaboration between the two. This means that there is still some room for improvement in pre-sales dialogue generation. And \acp{CRS} are still worth exploring because it can achieve comparable performance without spending a lot of resources and time for both training and inference.
\end{enumerate*}

\section{Conclusions and Future Work}
% - What did we do?
% - what did we find?
% - what does it mean / what are the implications?
% - what are the limitations?
% - what should we do next?
% In this paper, we explored the combination of conversational recommender system (CRS) and large language model (LLM) in E-commerce pre-sales dialogues. 
% Specifically, we proposed two types of collaboration, namely ``CRS assisting LLM'' and ``LLM assisting CRS''. 
% We explored the effectiveness of collaborations between two LLMs and two CRSs on four tasks of E-commerce pre-sales dialogues. 
% We conducted numerous experiments and performed careful analysis and findings. 
% We found that collaborations of CRS and LLM were remarkably effective in certain cases, providing insights for the research and application of E-commerce pre-sales dialogues. 
% It also inspired further exploration of collaboration between general large models and private small models in other domains and areas. 
% For future work, we planned to explore cross-category collaborations between CRS and LLM.

In this paper, we investigated the integration of conversational recommender systems (CRS) and large language models (LLM) in E-commerce pre-sales dialogues. 
Specifically, we proposed two collaboration strategies: ``CRS assisting LLM'' and ``LLM assisting CRS''. 
We evaluate the effectiveness of these collaborations between two LLMs and two CRSs on four tasks related to E-commerce pre-sales dialogues.
Through extensive experiments and careful analysis, we found that the collaboration between CRS and LLM can be highly effective in certain scenarios, providing valuable insights for both research and practical applications in E-commerce pre-sales dialogues. 
Additionally, our findings can inspire exploration of collaborations between general large models and private small models in other domains and areas. 

% For future work, we plan to explore cross-category collaborations between CRS and LLM.
% For future work, we plan to further explore the effectiveness of CRS and LLM i.e., two different strategies for generating responses, in collaborating on pre-sales dialogue generation.
% In addition, we plan to explore collaboration between LLMs and CRSs across categories. For example, a CRS under the Shoes category provides information to an LLM under the Fashion category to constitute a recommended combination, e.g., dress, pants, and shoes.

% For future work, we plan to further investigate the effectiveness of CRS and LLM, which are two distinct approaches for generating responses.
For future work, we plan to examine the collaboration between LLMs and CRSs across various categories. For instance, a CRS within the \textit{Shoes} category could provide information to a LLM in the \textit{Fashion} category, resulting in a recommended combination such as a dress, pants, and shoes.

% 页数（8页）
% ● Abstract：0.25
% ● introduction：1
% ● related work：0.5
% ● Method： 1.5
% ● experimental setup：1
% ● results：3
% ● Conclusion：0.25

% \clearpage
\clearpage
\section*{Limitations}

One limitation of our work is that our findings may only apply to \acfp{LLM} around 7B parameters.
% In this work, we adopted two \acp{}
% parameters of 7B and 6B for the large model, which are the more widely adopted sizes.
In this work, we select two \acp{LLM} that are widely used, namely ChatGLM-6B and Chinese-Alpaca-7B.
Related work reveals that there is some variation in the capability of \acp{LLM} with different parameter sizes~\cite{wei2022emergent}.
% For models with larger parameters, such as 100B, more GPU resources are needed to explore the effect of combining CRS and LLM.
For \acp{LLM} with more parameters, such as 100B, more GPU resources and time are needed to explore the effects of combining \acp{CRS} and \acp{LLM}.
In addition, \acp{LLM} are constantly being updated. Recently ChatGLM2-6B and Chinese-Alpaca2-13B have been open sourced.\footnote{\url{https://github.com/THUDM/ChatGLM2-6B} \\ \url{https://github.com/ymcui/Chinese-LLaMA-Alpaca-2}} 
They show better performance than ChatGLM-6B and Chinese-Alpaca-7B on multiple benchmarks, and may have higher results on E-commerce pre-sales dialogues as well.
However, we believe that the combination of \acp{LLM} and \acp{CRS} is still worth researching.
% Besides, we will explore whether similar findings exist in other domains of recommendations.

\section*{Ethics Statement}
% The dataset used for this work contains user information that may be of interest to user privacy. 
In this paper, we explore the effectiveness of combining LLM and CRS on e-commerce pre-sales conversations. 
We are committed to using and modify U-NEED dataset only for research purposes and not for commercial exploitation. 
Our proposed approach does not generate harmful content or raise ethical issues.
% We sent an application email to the owner of the dataset and then we are granted the right to use the dataset.
% Scientific work published at EMNLP 2023 must comply with the \href{https://www.aclweb.org/portal/content/acl-code-ethics}{ACL Ethics Policy}. We encourage all authors to include an explicit ethics statement on the broader impact of the work, or other ethical considerations after the conclusion but before the references. The ethics statement will not count toward the page limit (8 pages for long, 4 pages for short papers).

\section*{Acknowledgments}
We thank the anonymous reviewers for their helpful comments. 
% This work is supported by the Science and Technology Innovation 2030 Major Project of China (No. 2020AAA0108605) and National Natural Science Foundation of China (No. 62076081, No. 61772153, and No. 61936010).
% This work is supported by the National Key Research and  Development Program (No. 2022YFF0902100) and National Natural Science Foundation of China (No. 62076081 and No. 61936010).
This work is supported by the National Key Research and Development Program (No. 2022YFF0902100), National Natural Science Foundation of China (No. 62076081 and No. 61936010), and Nature Scientific Foundation of Heilongjiang Province (No. YQ2021F006).

% Entries for the entire Anthology, followed by custom entries
\bibliography{reference}
\bibliographystyle{acl_natbib}

\appendix

\section{Pre-sales Dialogue Tasks}
\label{sec:app_task_formulation}
To evaluate the performance of \acp{CRS}, \acp{LLM} and their collaborations in E-commerce scenarios, we adopt four challenging tasks proposed in U-NEED dataset~\cite{liu2023u}. 
The four challenging tasks are: 
\begin{enumerate*}[label=(\roman*)]
    \item pre-sales dialogue understanding
    \item user needs elicitation
    \item user needs-based recommendation and
    \item pre-sales dialogue generation.
\end{enumerate*}

Pre-sales dialogue understanding aims to understand utterances of both users and customer service staff.
First, identify the attributes that are related to products.
And second, extract preferences related to the identified attributes.
For example, when a user says, ``Which of your thermal underwear is the warmest? Recommend one?''
This task aims to obtain semantic frames \{ (``Functional requirement'', ``Warmest''), (``Category'', ``Thermal underwear'') \}, where ``Functional requirement'' and ``Category'' are attributes related to products.
``Warmest'' and ``Thermal underwear'' are preferences.

User needs elicitation aims to select attributes that can elicit more information about user needs.
The inputs for this task are the dialogue context and the identified user needs, i.e., \{ (``Functional requirement'', ``Warmest''), (``Category'', ``Thermal underwear'') \}.
The output is a set of attributes, e.g., \{``Category'', ``Price''\}.

User needs-based recommendation aims to recommend products that satisfy explicit and implicit user needs.
Explicit user needs refer to the needs and preferences expressed by the user in the ongoing dialogue, i.e., \{ (``Functional requirement'', ``Warmest''), (``Category'', ``Thermal underwear'') \}.
Implicit user needs are related to user behaviors outside of the pre-sales dialogue. 
Users usually view some items before starting a dialogue with the customer service staff. 
In addition, they browse through items while talking to customer service staff.
Such behaviors can reflect implicit user needs to some extent.
The inputs to this task are explicit and implicit user needs, i.e., identified semantic frames and user behaviors, and the output is a collection of items.

Pre-sales dialogue generation aims to generate a response based on given information about user needs.
Information consists of a collection of attributes and a collection of items.
The collection of attributes is the output of the user needs elicitation task, i.e., the attributes that may elicit more information about the user's needs.
The collection of items is the output of the user needs-based recommendation task, i.e., items that satisfy the current user needs.
The inputs to this task are the dialogue context, the collection of attributes and the collection of items.
The output is a response, which may be a query asking a question about an attribute, e.g.,``What are your requirements for the thermal underwear?'' or a recommendation reason for recommending an item, e.g., ``Recommended item: 655398643290. This one is seamless and made of double-sided fleece. You can take a look.''

\section{Baselines for Pre-sales Dialogue Tasks}
\label{sec:app_baselines}

Following~\citet{liu2023u}, we 
\begin{enumerate*}[label=(\roman*)]
    \item select Bert~\cite{Devlin2019bert}, Bert+CRF~\cite{Souza2019bertcrf} and Bert+BiLSTM+CRF~\cite{Dai2019bertbicrf} as baselines for the pre-sales dialogue understanding task; 
    \item select DiaMultiClass~\cite{li-etal-2020-rethinking} and DiaSeq~\cite{li-etal-2020-rethinking} as baselines for the user needs elicitation task; 
    \item select Bert~\cite{Devlin2019bert}, SASRec~\cite{Kang2018sasrec} and TG-CRS~\cite{zhou2020towards} as baselines for the user needs-based recommendation task; 
    \item and select GPT-2~\cite{radford2019language} and KBRD~\cite{chen-etal-2019-towards} as baselines for the pre-sales dialogue generation task.
\end{enumerate*}

For the pre-sales dialogue understanding task, Bert, Bert+CRF, and Bert+BiLSTM+CRF adopt the sequence labeling approach to identify the semantic frames.
Bert considers only the representation of the input utterance.
Bert+CRF takes into account the sequential relationships of the predicted tags in addition to the representation of the input utterance. 
Whereas Bert+BiLSTM+CRF adds bidirectional information encoding after obtaining the representation of the input and considers the sequential relationships of the predicted tags to compute the probability of each tag.

For the user needs elicitation task, DiaMultiClass and DiaSeq employ a multi-label classification approach to determine the collection of attributes.DiaMultiClass computes the probability of each attribute based on the representation of the inputs.DiaSeq computes the probability of each attribute based on the sequential relationship between semantic frames.
% Bert~\cite{Devlin2019bert} calculates the probability of each tag based on the representation of the input utterance.
% Bert+CRF~\cite{Souza2019bertcrf} calculates the probability of each tag considering the representation of the input utterance as well as the sequential relationship between predicted tags.
% Bert+BiLSTM+CRF~\cite{Dai2019bertbicrf} considers bi-directional information after the encoding of bert to obtain a better representation of the input utterance. It also considers the sequential relationship between predicted tags. 
% DiaMultiClass~\cite{li-etal-2020-rethinking}] encodes the input text and then predicts the probability of each attribute. In the inference stage, we set a threshold of 0.5 to output the prediction results.
% DiaSeq~\cite{li-etal-2020-rethinking}] utilizes a GRU-based decoder to predict attribute combinations. 

The baseline for the user needs-based recommendation task is Bert, SASRec and TG-CRS.
Bert calculates the probability of an item based on the representation of the input.
SASRec calculates the probability of an item based on the sequential relationships of user behaviors.
TG-CRS considers both dialogue context and sequential user behaviors.
% Bert~\cite{Devlin2019bert}] encodes the dialogue context and then makes recommendations.
% SASRec~\cite{Kang2018sasrec}] employs a self-attention mechanism to capture long-term semantics in users' historical behaviors to generate recommendations.
% TG-CRS~\cite{zhou2020towards}] fuses the representation of dialogue context and users' historical behaviors to make recommendations. 

For the pre-sales dialogue generation task, the baselines are GPT-2 and KBRD.
GPT-2 is a commonly used pre-trained language model for the dialogue generation task.
KBRD utilizes a switching mechanism to introduce tokens related to the recommended items during the decoding of responses.

% GPT-2~\cite{radford2019language}] is a pre-training text generation model and fine-tuned on \ac{U-NEED} dataset.
% KBRD~\cite{chen-etal-2019-towards}] applies a transformer with enhanced modeling of word weight based on knowledge graphs.;

\section{Combinations of \acp{LLM} and \acp{CRS}}
\label{sec:app_variants}

We define four variants of \ac{LLM} assists \ac{CRS}:
\begin{itemize}
    \item {CLLM-BCRS} refers that ChatGLM assists BART-based CRS.
    \item {CLLM-CCRS} refers that ChatGLM assists CPT-based CRS.
    \item {ALLM-BCRS} refers that Chinese-Alpaca assists BART-based CRS.
    \item {ALLM-CCRS} refers that Chinese-Alpaca assists BART-based CRS.
\end{itemize}

We define four variants of \ac{CRS} assists \ac{LLM}:
\begin{itemize}
    \item {BCRS-CLLM} refers that BART-based CRS assists ChatGLM.
    \item {CCRS-CLLM} refers that CPT-based CRS assists ChatGLM.
    \item {BCRS-ALLM} refers that BART-based CRS assists Chinese-Alpaca.
    \item {CCRS-ALLM} refers that CPT-based CRS assists Chinese-Alpaca.
\end{itemize}

\section{Evaluation Metrics}
\label{sec:app_evaluation_metrics}

Following~\citet{liu2023u}, for the pre-sales dialogue understanding and user needs elicitation tasks, we set Precision, Recall and F1 score as evaluation metrics.
Precision is the proportion of correctly selected tags to the total number of selected tags.
Recall is the ratio of correctly selected tags to the original number of correct tags.
The F1 score is calculated by taking the harmonic mean of precision and recall.

Regarding the user needs-based recommendation task, the evaluation metrics in U-NEED~\cite{liu2023u} are Hit@10, Hit@50 and MRR@50. Due to the limitation of the input length of LLMs, where each product contains attributes and attribute values, we can provide a maximum of 20 candidate products. Therefore, in order to compare whether the collaborative approach improves the performance of CRSs, we measure Accuracy (Hit@1), Hit@5 and MRR@5.
The Hit@K metric represents the proportion of relevant items that are present in the top-K results out of all the relevant items.
The MRR@K score is determined by taking the average of the reciprocal ranks of the top-K items in a ranking. If an item does not appear in the top-K positions, its reciprocal rank is set to 0.

The evaluation metrics for the pre-sales dialogue generation task are Distinct@1, Informativeness and Relevance.
Distinct@1 is computed as the average of the fraction of distinct 1-grams out of all 1-grams in a response.
Distinct@1 measures the diversity of generated responses.
Informativeness and relevance are for human evaluation. We randomly sample 100 dialogues and we recruit 12 annotators to evaluate 1400 responses from 14 methods on these 100 dialogues.
Informativeness is calculated as the average informativeness of all generated responses.
Relevance is determined as the average relevance degree of all generated responses.
The annotators evaluate the extent to which a generated response includes information about the product, as compared to the ground truth.
The score of informativeness and relevance ranges from 1 to 5, and we calculate the average score from all annotators to obtain the final score.
% We calculate the Fleiss’s kappa~\cite{fleiss1971measuring} to measure the inter-annotator agreements.

\section{Implementation Details}

% \todo{
% We provide the code used in the experiments.
% As the dataset consists of user privacy, we are not permitted to distribute it.
% The demo data is in the same json format as the full data, so you can reproduce our experiments using the code we provide, as long as you get the real data.
We implement \acp{CRS} based on UniMIND.\footnote{\url{https://github.com/dengyang17/UniMIND}}
The code is available online.\footnote{\url{https://github.com/LeeeeoLiu/LLM-CRS}}
% We use the a100-sxm-80gb and a100-sxm4-80gb GPUs for fine-tuning the LLM and training the CRS.
% Due to the limitation of input length, the number of candidate products for the user needs-based recommendation task is set to 20.
% For training CRS, we train 15 epochs in the first stage and 20 epochs in the second stage.
% When fine-tuning LLM, we fine-tune 10 epochs using LoRA.
% We utilize the a100-sxm-80gb and a100-sxm4-80gb GPUs to fine-tune the LLM and train the CRS.
% Due to input length restrictions, we limit the number of candidate products to 20 for the user needs-based recommendation task.
% For CRS training, we conduct 15 epochs in the initial stage and 20 epochs in the subsequent stage.
% During the fine-tuning of the LLM, we perform 10 epochs of fine-tuning.
For CRSs, we use a NVIDIA A100-SXM4-80GB gpu and train model for 10 epochs, with a duration of approximately 12 hours. For LLMs, we use a NVIDIA A100-SXM4-80GB gpu and train model for 3 epochs, with a duration of approximately 9 hours. 
%  using LoRA.
% }

\section{Examples of Fine-tuning LLMs}
\label{sec:app_sft_examples}

\begin{table*}[!t]
    \caption{Three examples of fine-tuning \acp{LLM} for pre-sales dialogue understanding task in Appliance, Beauty and Fashion categories.}
    \resizebox{\textwidth}{!}{
    \begin{tabular}{cp{5.2cm}p{8.2cm}}
    \toprule
     & \textbf{Example} & \textbf{Translation} \\ 
    \midrule 
    Instruction & \begin{CJK}{UTF8}{gbsn} 结合大家电行业售前对话，识别当前用户或者客服输入中涉及的商品相关的属性值和对应的属性。针对用户输入，需要识别出属性和属性值。客服输入中属性值可能为空。\end{CJK} & Combined with the pre-sales dialogue in Appliance category, identify the product-related attribute values and corresponding attributes involved in the input of the current user or customer service. For user input, attributes and attribute values need to be identified. The attribute value in the customer service input may be empty.
    \\  
    Input& \begin{CJK}{UTF8}{gbsn} 售前对话：用户：帮我推荐一款普通洗衣机，性价比高，皮实耐用的，不要烘干功能的   当前输入：客服：波轮还是滚筒呢\end{CJK} & Pre-sale dialogue: User: Help me recommend an ordinary washing machine with high cost performance, durable leather, and no drying function   Current input: Customer service: Wave wheel or drum    \\ 
    Output& \begin{CJK}{UTF8}{gbsn} 洗衣机类型：波轮;洗衣机类型：滚筒 \end{CJK} & washing machine type: wave wheel; washing machine type: drum \\
    \midrule
    \midrule
    Instruction & \begin{CJK}{UTF8}{gbsn} 结合美妆行业售前对话，识别当前用户或者客服输入中涉及的商品相关的属性值和对应的属性。针对用户输入，需要识别出属性和属性值。客服输入中属性值可能为空。\end{CJK} & Combined with the pre-sales dialogue in Beauty category, identify the product-related attribute values and corresponding attributes involved in the input of the current user or customer service. For user input, attributes and attribute values need to be identified. The attribute value in the customer service input may be empty.
    \\  
    Input& \begin{CJK}{UTF8}{gbsn} 售前对话：用户：你们店有没有套装[SEP]客服：您想要什么类型的呢[SEP]用户：补水[SEP]客服：水乳吗亲亲[SEP]用户：嗯，需要水乳[SEP]客服：亲亲需要祛痘的吗[SEP]用户：需要祛痘的   当前输入：客服：亲亲是想解决红肿痘痘还是闭口的么    \end{CJK} & Pre-sale dialogue: User: Do you have a set in your store? [SEP] Customer service: What type do you want? [SEP] User: Hydration [SEP] Customer service: Do you want milk? Kiss [SEP] User: Well, I need water Milk[SEP]Customer service: Kiss, do you need to get rid of acne[SEP]User: Need to get rid of acne Current input: Customer service: Do you want to solve red, swollen, pimples or keep your mouth shut?    \\ 
    Output& \begin{CJK}{UTF8}{gbsn} 肌肤问题：红肿痘痘;肌肤问题：闭口 \end{CJK} & skin problem: redness, swelling and acne; skin problem: shut up \\
    \midrule
    \midrule
    Instruction & \begin{CJK}{UTF8}{gbsn} 结合鞋类行业售前对话，识别当前用户或者客服输入中涉及的商品相关的属性值和对应的属性。针对用户输入，需要识别出属性和属性值。客服输入中属性值可能为空。\end{CJK} & Combined with the pre-sales dialogue in Fashion category, identify the product-related attribute values and corresponding attributes involved in the input of the current user or customer service. For user input, attributes and attribute values need to be identified. The attribute value in the customer service input may be empty.
    \\  
    Input& \begin{CJK}{UTF8}{gbsn} 售前对话：用户：还有没有其他款推荐[SEP]客服：要什么材质的呢亲爱哒[SEP]用户：就是冰丝的那种[SEP]客服：仅发送商品链接[SEP]客服：这个是莫代尔材质冰丝触感的[SEP]客服：仅发送商品链接[SEP]客服：这个是冰丝材质的[SEP]用户：冬天了穿哪款好点 当前输入：用户：我儿子就喜欢穿金利来这个牌子的 \end{CJK} & Pre-sale dialogue: User: Do you have any other recommendations? [SEP] Customer service: What material do you want, dear? [SEP] User: It’s the ice silk one [SEP] Customer service: Only send product links [SEP] Customer service : This is made of modal material with ice silk touch [SEP] Customer service: Only send product links [SEP] Customer service: This is made of ice silk [SEP] User: Which one is better to wear in winter Current input: User: My son I like to wear the brand of Jinlilai.    \\ 
    Output& \begin{CJK}{UTF8}{gbsn} 人群：儿子;品牌：金利来 \end{CJK} & Crowd: Son; Brand: Goldlion \\
    \bottomrule
    \end{tabular}
    }
    \label{tab:sft_example_task1_1}
\end{table*}

\begin{table*}[!t]
    \caption{Three examples of fine-tuning \acp{LLM} for user needs elicitation task in Appliance, Beauty and Fashion categories.}
    \resizebox{\textwidth}{!}{
    \begin{tabular}{cp{5.2cm}p{8.2cm}}
    \toprule
    % & \multicolumn{3}{c}{Beauty} & \multicolumn{3}{c}{Shoes} & \multicolumn{3}{c}{Phones} & \multicolumn{3}{c}{All 5 categories}  \\ \cmidrule(r){2-4}\cmidrule(r){5-7}\cmidrule(r){8-10} \cmidrule{11-13}
    % Methods & P & R & F1 & P & R & F1 & P & R & F1 & P & R & F1 \\ \midrule
     & \textbf{Example} & \textbf{Translation} \\ 
    \midrule 
    Instruction & \begin{CJK}{UTF8}{gbsn} 依据大家电行业售前对话，选择一系列的属性，来引导用户提供更多关于需求的偏好信息。结果中可以包含属性值，也可以不包含属性值。\end{CJK} & According to the pre-sales dialogue in Appliance category, select a series of attributes to guide users to provide more preference information about needs. Attribute values may or may not be included in the result.
    \\  
    Input& \begin{CJK}{UTF8}{gbsn} 售前对话：用户：帮我推荐一款普通洗衣机，性价比高，皮实耐用的，不要烘干功能的[SEP]客服：波轮还是滚筒呢[SEP]用户：滚筒的[SEP]用户：功能简单的[SEP]客服：仅发送商品链接[SEP]客服：仅发送商品链接[SEP]用户：有小天鹅的吗[SEP]客服：(1)专属净柔洗程序，柔和洗护爱衣，独特的全方位按摩，如同手洗般轻柔、揉搓间为衣物重塑洁净与柔软；(2)95度高温煮洗，扫净藏于衣物纤维中的病毒细菌，长效杀菌灭毒，99.9\%健康除菌(3)wifi手机远程控制，随时随地，想穿就穿(4)特色羽绒服洗，分多段进水，洗涤节拍柔和，预防羽绒服漂浮水面或破损，洗护均匀，贴心呵护(5)BLDC变频电机，脱水更快更彻底，洁净少残留[SEP]用户：波轮的哪款性价比高？皮实耐用 \end{CJK} & Pre-sale conversation: User: Help me recommend an ordinary washing machine with high cost performance, durable leather, and no drying function [SEP] Customer service: Wave wheel or drum [SEP] User: Drum [SEP] User: Simple function [SEP] Customer service: Only send product links [SEP] Customer service: Only send product links [SEP] User: Do you have Little Swan? Azimuth massage, as gentle as washing by hand, reshape the cleanliness and softness of the clothes between rubbing; (2)Boil and wash at 95 degrees high temperature, sweep away the viruses and bacteria hidden in the fibers of the clothes, long-term sterilization and disinfection, 99.9\% healthy sterilization (3)Wifi mobile phone remote control , anytime, anywhere, you can wear it as you want (4)Wash the special down jacket, enter the water in multiple stages, the washing cycle is soft, prevent the down jacket from floating on the water or damage, even washing and care, caring (5)BLDC inverter motor, dehydration is faster and more thorough, clean and less residue [SEP ] User: Which one of the wave wheel is more cost-effective? Durable
    \\ 
    Output& \begin{CJK}{UTF8}{gbsn} 价位 \end{CJK} & Price \\
    \midrule
    \midrule
    Instruction & \begin{CJK}{UTF8}{gbsn} 依据美妆行业售前对话，选择一系列的属性，来引导用户提供更多关于需求的偏好信息。结果中可以包含属性值，也可以不包含属性值。\end{CJK} & According to the pre-sales dialogue in Beauty category, select a series of attributes to guide users to provide more preference information about needs. Attribute values may or may not be included in the result.    \\  
    Input& \begin{CJK}{UTF8}{gbsn} 售前对话：用户：你们店有没有套装    \end{CJK} & Pre-sale dialogue: User: Do you have any suits in your store?   \\ 
    Output& \begin{CJK}{UTF8}{gbsn} 功效 \end{CJK} & Efficacy \\
    \midrule
    \midrule
    Instruction & \begin{CJK}{UTF8}{gbsn} 依据服装行业售前对话，选择一系列的属性，来引导用户提供更多关于需求的偏好信息。结果中可以包含属性值，也可以不包含属性值。\end{CJK} & According to the pre-sales dialogue in Fashion category, select a series of attributes to guide users to provide more preference information about needs. Attribute values may or may not be included in the result.    \\  
    Input& \begin{CJK}{UTF8}{gbsn} 售前对话：用户：还有没有其他款推荐 \end{CJK} & Pre-sale dialogue: User: Do you have any other recommendations?
    \\ 
    Output& \begin{CJK}{UTF8}{gbsn} 材质 \end{CJK} & Material \\
    \bottomrule
    \end{tabular}
    }
    \label{tab:sft_example_task2_1}
\end{table*}

\begin{table*}[!t]
    \caption{An example of fine-tuning \acp{LLM} for user needs-based recommendation task in Fashion category.}
    \resizebox{1\textwidth}{!}{
    \begin{tabular}{cp{6cm}p{8cm}}
    \toprule
     & \textbf{Example} & \textbf{Translation} \\ 
    \midrule 
    Instruction & \begin{CJK}{UTF8}{gbsn} 根据服装行业售前对话中用户表达的需求和偏好信息以及候选商品信息，从候选商品A-T中选择最有可能满足用户需求、偏好的商品推荐给用户。\end{CJK} & According to the demand and preference information expressed by the user in the pre-sales dialogue in Fashion category and the candidate product information, the product that is most likely to meet the user's needs and preferences is selected from the candidate products A-T and recommended to the user. \\  
    Input& \begin{CJK}{UTF8}{gbsn} 售前对话：用户：还有没有其他款推荐[SEP]客服：要什么材质的呢亲爱哒[SEP]用户：就是冰丝的那种 各候选商品对应的属性和属性值：A的价格区间是高，功能需求是舒适，季节是夏，性别是男，服装厚度是薄款，材质是冰丝、棉质、莫代尔，款式是平角、无痕、简单，类目是男平角内裤[SEP]B的价格区间是高[SEP]C的价格区间是高[SEP]D的价格区间是高，性别是男，类目是睡衣/家居服套装[SEP]E的价格区间是高[SEP]F的价格区间是中，功能需求是保暖，季节是秋，性别是女，类目是保暖套装[SEP]G的价格区间是高[SEP]H的价格区间是高[SEP]I的价格区间是高，功能需求是保暖，性别是女，服装厚度是薄款，类目是保暖套装[SEP]J的价格区间是高[SEP]K的价格区间是高[SEP]L的价格区间是高[SEP]M的价格区间是高[SEP]N的价格区间是高[SEP]O的价格区间是高[SEP]P的价格区间是高[SEP]Q的价格区间是高[SEP]R的价格区间是高[SEP]S的价格区间是高[SEP]T的价格区间是高，款式是v领，类目是保暖套装    \end{CJK} & Pre-sale dialogue: User: Do you have any other recommendations [SEP] Customer service: What material do you want? Dear [SEP] User: It is the kind of ice silk The attributes and attribute values   corresponding to each candidate product: A The price range is high, the functional requirement is comfortable, the season is summer, the gender is male, the clothing thickness is thin, the material is ice silk, cotton, modal, the style is boxer, no trace, simple, and the category is men's boxer underwear [SEP] The price range of B is high [SEP] The price range of C is high [SEP] The price range of D is high, the gender is male, and the category is pajamas/home service sets [SEP] The price range of E is high [ The price range of SEP]F is medium, the functional requirement is to keep warm, the season is autumn, the gender is female, and the category is thermal suits. The price range of [SEP]G is high. The price range of [SEP]H is high.[SEP]I The price range is high, the functional requirement is to keep warm, the gender is female, the clothing thickness is thin, and the category is thermal suit [SEP]J, the price range is high[SEP]K, the price range is high[SEP]L It is high [SEP] the price range of M is high [SEP] the price range of N is high [SEP] the price range of O is high [SEP] the price range of P is high [SEP] the price range of Q is high [SEP] The price range of R is high [SEP] the price range of S is high [SEP] the price range of T is high, the style is v-neck, and the category is thermal suit  \\ 
    Output& \begin{CJK}{UTF8}{gbsn}A\end{CJK} & A \\
    \bottomrule
    \end{tabular}
    }
    \label{tab:sft_example_task3_1}
\end{table*}

\begin{table*}[!t]
    \caption{Three examples of fine-tuning \acp{LLM} for pre-sales dialogue generation task in Appliance, Beauty and Fashion categories.}
    \resizebox{\textwidth}{!}{
    \begin{tabular}{cp{6.8cm}p{8.8cm}}
    \toprule
    % & \multicolumn{3}{c}{Beauty} & \multicolumn{3}{c}{Shoes} & \multicolumn{3}{c}{Phones} & \multicolumn{3}{c}{All 5 categories}  \\ \cmidrule(r){2-4}\cmidrule(r){5-7}\cmidrule(r){8-10} \cmidrule{11-13}
    % Methods & P & R & F1 & P & R & F1 & P & R & F1 & P & R & F1 \\ \midrule
     & \textbf{Example} & \textbf{Translation} \\ 
    \midrule 
    Instruction & \begin{CJK}{UTF8}{gbsn} 根据大家电行业售前对话中已获取的信息、引导用户需求的属性、满足用户需求的商品信息，生成回应用户需求且用户容易理解的通俗回复。\end{CJK} & According to the information obtained in the pre-sales dialogue in Appliance category, the attributes that guide the user's needs, and the product information that meets the user's needs, generate a popular reply that responds to the user's needs and is easy for the user to understand. \\  
    Input& \begin{CJK}{UTF8}{gbsn} 售前对话：用户：帮我推荐一款普通洗衣机，性价比高，皮实耐用的，不要烘干功能的[SEP]客服：波轮还是滚筒呢[SEP]用户：滚筒的[SEP]用户：功能简单的[SEP]客服：仅发送商品链接[SEP]客服：仅发送商品链接[SEP]用户：有小天鹅的吗[SEP]客服：(1)专属净柔洗程序，柔和洗护爱衣，独特的全方位按摩，如同手洗般轻柔、揉搓间为衣物重塑洁净与柔软；(2)95度高温煮洗，扫净藏于衣物纤维中的病毒细菌，长效杀菌灭毒，99.9\%健康除菌(3)wifi手机远程控制，随时随地，想穿就穿(4)特色羽绒服洗，分多段进水，洗涤节拍柔和，预防羽绒服漂浮水面或破损，洗护均匀，贴心呵护(5)BLDC变频电机，脱水更快更彻底，洁净少残留[SEP]用户：波轮的哪款性价比高？皮实耐用 已获取的用户需求偏好信息：品类：洗衣机、功能需求：皮实耐用、不要烘干、功能简单、皮实耐用、款式：性价比高、性价比高、洗衣机类型：滚筒、波轮、品牌：小天鹅 引导用户需求的属性：价位：
    \end{CJK} & Pre-sale conversation: User: Help me recommend an ordinary washing machine with high cost performance, durable leather, and no drying function [SEP] Customer service: Wave wheel or drum [SEP] User: Drum [SEP] User: [SEP] Customer service with simple functions: only send product links [SEP] customer service: only send product links [SEP] user: do you have Little Swan? The all-round massage is as gentle as hand washing, and the rubbing will reshape the cleanliness and softness of the clothes; (2)Boil and wash at 95 degrees high temperature, sweep away the virus bacteria hidden in the fibers of the clothes, long-term sterilization and disinfection, 99.9\% healthy sterilization (3)wifi mobile phone Remote control, you can wear it anytime, anywhere (4)Special down jacket washing, multi-stage water intake, gentle washing cycle, prevent down jacket from floating or damaged, even washing and care, caring (5)BLDC frequency conversion motor, dehydration is faster and more thorough, clean and less residue [SEP] User: Which one of the wave wheel is more cost-effective? Durable Leather Acquired user demand preference information: category: washing machine, functional requirements: durable leather, no drying, simple function, durable leather, style: cost-effective, cost-effective, washing machine type: drum, pulsator, brand: Little Swan Attributes to guide user needs: Price: \\ 
    Output& \begin{CJK}{UTF8}{gbsn} 预算多少呢亲 \end{CJK} & How much is the budget? \\
    \midrule
    \midrule
    Instruction & \begin{CJK}{UTF8}{gbsn} 根据美妆行业售前对话中已获取的信息、引导用户需求的属性、满足用户需求的商品信息，生成回应用户需求且用户容易理解的通俗回复。\end{CJK} & According to the information obtained in the pre-sales dialogue in Beauty category, the attributes that guide the user's needs, and the product information that meets the user's needs, generate a popular reply that responds to the user's needs and is easy for the user to understand. \\   
    Input& \begin{CJK}{UTF8}{gbsn} 售前对话：用户：你们店有没有套装 已获取的用户需求偏好信息：品类：套装 引导用户需求的属性：功效： \end{CJK} & Pre-sales dialogue: User: Do you have suits in your store  Acquired user demand preference information: Category: Set  Attributes to guide user needs: Function:    \\ 
    Output& \begin{CJK}{UTF8}{gbsn} 您想要什么类型的呢 \end{CJK} & what type do you want \\
    \midrule
    \midrule
    Instruction & \begin{CJK}{UTF8}{gbsn} 根据服装行业售前对话中已获取的信息、引导用户需求的属性、满足用户需求的商品信息，生成回应用户需求且用户容易理解的通俗回复。\end{CJK} & According to the information obtained in the pre-sales dialogue in 
Fashion category, the attributes that guide the user's needs, and the product information that meets the user's needs, generate a popular reply that responds to the user's needs and is easy for the user to understand. \\  
    Input& \begin{CJK}{UTF8}{gbsn} 售前对话：用户：还有没有其他款推荐 已获取的用户需求偏好信息：款式： 引导用户需求的属性：材质： \end{CJK} & Pre-sale dialogue: User: Do you have any other recommendations?   Obtained user demand preference information: Style:   Attributes that guide user demand: Material: \\ 
    Output& \begin{CJK}{UTF8}{gbsn} 要什么材质的呢亲爱哒 \end{CJK} & What material do you want dear? \\
    \bottomrule
    \end{tabular}
    }
    \label{tab:sft_example_task4_1}
\end{table*}

We give examples of the instructions, inputs, and outputs used to fine-tune the LLMs for each task in Tables \ref{tab:sft_example_task1_1}, \ref{tab:sft_example_task2_1}, \ref{tab:sft_example_task3_1}, and \ref{tab:sft_example_task4_1}, respectively.

\section{Examples of Collaborations of CRSs and LLMs}

\begin{figure*}[!t]
    \centering
    \includegraphics[width=1\textwidth]{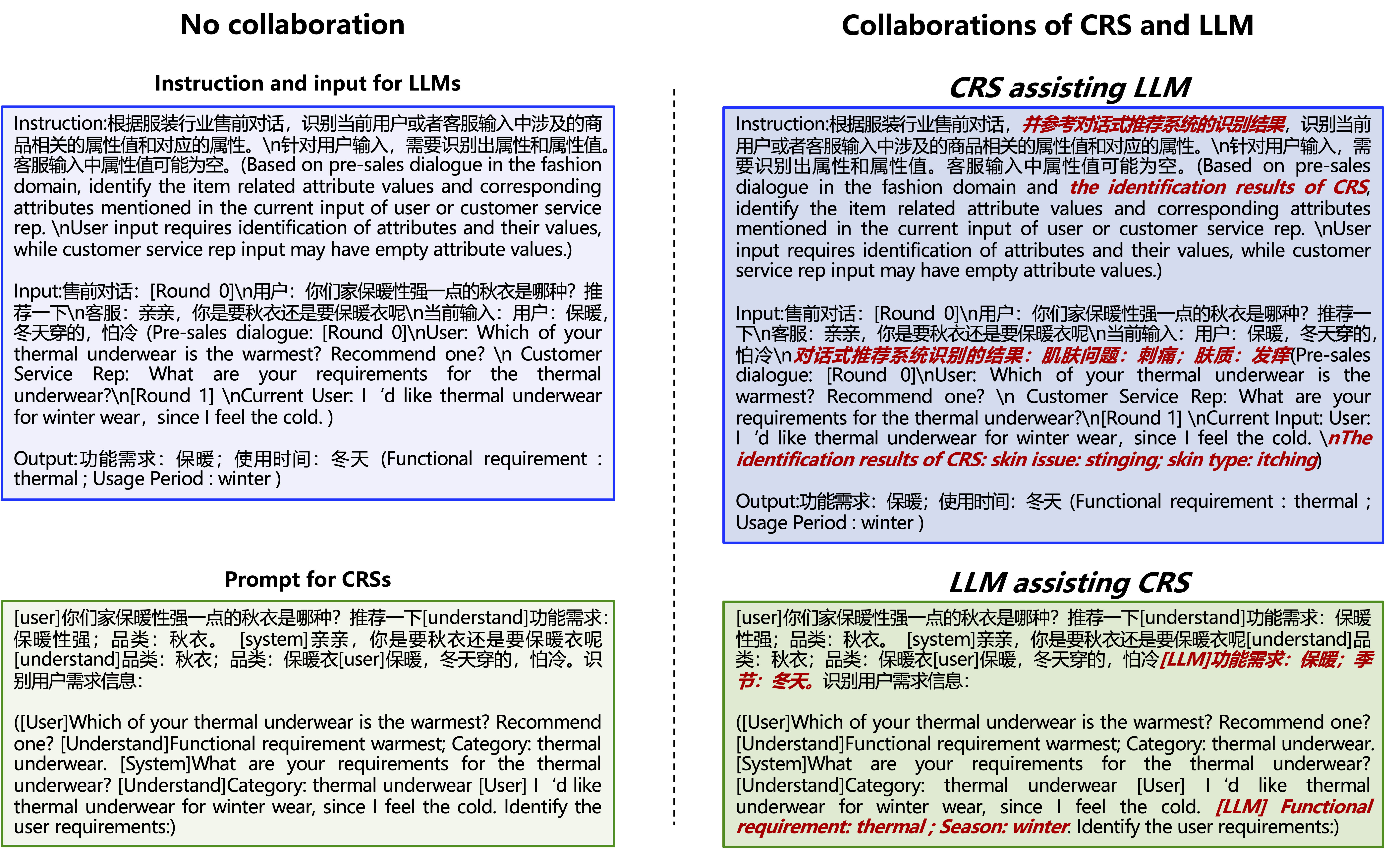}
    \caption{An example of collaboration between CRS and LLM on the pre-sales dialogue understanding task. Left side displays data used to fine-tune a LLM and train a CRS independently. The right side shows two cases of combining the two. Collaboration content is highlighted in red italics.}
    \label{fig:app_ex_1}
\end{figure*}

\begin{figure*}[!t]
    \centering
    \includegraphics[width=1\textwidth]{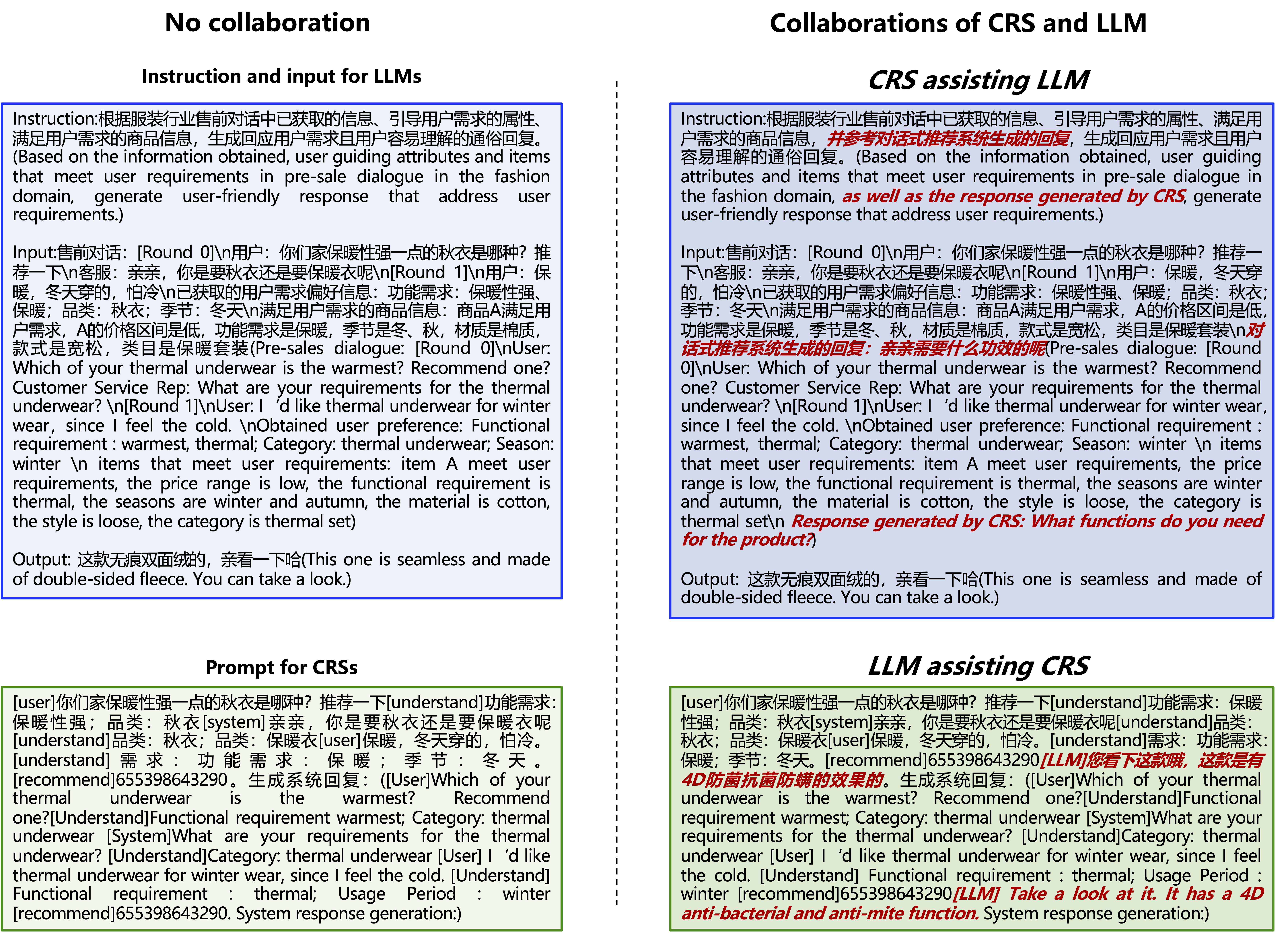}
    \caption{An example of collaboration between CRS and LLM on the pre-sales dialogue generation task. Left side displays data used to fine-tune a LLM and train a CRS independently. The right side shows two cases of combining the two. Collaboration content is highlighted in red italics.}
    \label{fig:app_ex_4}
\end{figure*}

We show examples of inputs (instructions) for collaborations between CRSs and LLMs on pre-sales dialogue understanding and generation tasks in Fig. \ref{fig:app_ex_1} and Fig. \ref{fig:app_ex_4}, respectively.

% \begin{figure*}[!t]
%     \centering
%     \includegraphics[width=1\textwidth]{pics/app_ex_2.png}
%     \caption{An illustration of pre-sales dialogue tasks and fine-tuning data.}
%     \label{fig:app_ex_2}
% \end{figure*}

\end{document}